\documentclass[twocolumn, a4paper,11pt]{article}

\usepackage{geometry}
\usepackage{titling}
\usepackage[utf8]{inputenc}
\DeclareUnicodeCharacter{2212}{-}

\usepackage{setspace}
\usepackage{multicol}
\usepackage{multirow}
\usepackage{afterpage}
\usepackage{rotating}
\usepackage{lineno}
\usepackage{xtab}
\usepackage{cuted}

\usepackage{amsmath}
\usepackage{amssymb}
\usepackage{microtype}
\usepackage{siunitx}
\usepackage[blocks]{authblk}
\usepackage{lscape}
\usepackage{gensymb}

\DeclareUnicodeCharacter{2713}{\checka}

\usepackage{dirtytalk}
\usepackage{txfonts}
\usepackage{soul}
\usepackage{tcolorbox}
\usepackage{enumitem}
\setlist[itemize]{leftmargin=*}
\usepackage{booktabs} 

\usepackage[table]{xcolor} 

\usepackage[numbers,square,sort&compress]{natbib}

\usepackage{graphicx}
\usepackage{float}
\usepackage{subcaption}
\usepackage[font={footnotesize},labelfont=bf]{caption}

\usepackage{blindtext}
\usepackage{textcomp}
\usepackage{capt-of} 

\usepackage{xr}
\usepackage[hidelinks]{hyperref}
\hypersetup{
    colorlinks=true,
    linkcolor=black, 
    filecolor=magenta,      
    urlcolor=cyan,
    }

\usepackage[capitalize]{cleveref}
\crefname{section}{Sec.}{Secs.}
\Crefname{section}{Section}{Sections}
\crefname{table}{Tab.}{Tabs.}
\Crefname{table}{Table}{Tables}

\usepackage{algorithm}
\usepackage[noend]{algpseudocode}

\renewcommand*{\thefootnote}{\roman{fnsymbol}}

\algnewcommand\algorithmicinput{\textbf{Preprocessing:}}
\algnewcommand\Preprocessing{\item[\algorithmicinput]}

\usepackage{styles_and_packages/arxiv}

\usepackage{printlen}

\newcommand{\rebuttal}[1]{\textcolor{black}{#1}}

\begin{document}

\title{What DINO saw: ALiBi positional encoding reduces positional bias in Vision Transformers}

\author[1]{{Moritz Pawlowsky}}

\author[2, 3]{{Antonis Vamvakeros}}

\author[1]{Alexander Weiss}

\author[1, 4]{Anja Bielefeld}

\author[2]{\\ {Samuel J. Cooper}\textsuperscript{*}  }

\author[2, 5]{Ronan Docherty}

\affil[1]{{\textit{\footnotesize Center for Materials Research, Justus Liebig University, 35392 Giessen}}}

\affil[2]{{\textit{\footnotesize Dyson School of Design Engineering, Imperial College London, London}}}

\affil[3]{{\textit{\footnotesize Finden ltd, Building R71, Rutherford Appleton Laboratory, Harwell Science and Innovation Campus}}}

\affil[4]{\textit{\footnotesize Institute for Sustainable Chemistry, Technische Universität Braunschweig}}

\affil[5]{{\textit{\footnotesize Department of Materials, Imperial College London, London}}}

\par\mbox{}

\begin{strip}
    \maketitle
    \vspace{2em}
    \centering
    \captionsetup{type=figure}
    \includegraphics{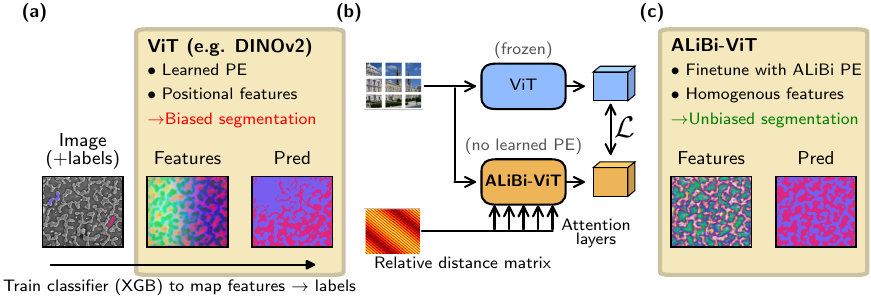}
    \captionof{figure}{Summary of our contribution. \textbf{(a)} \rebuttal{The choice of positional encoding (PE) for Vision Transformers (ViTs) - such as the DINO series - can} lead to positionally-biased features, which causes poor segmentations when used in zero-shot segmentation of out-of-distribution images. \textbf{(b)} We remove the \rebuttal{existing} PEs of \rebuttal{trained ViTs}, add 2D ALiBi PEs (based on relative token distances at each attention layer) and finetune to target the original embeddings. \textbf{(c)} This produces \rebuttal{models} with more homogenous features and better resulting segmentations.}
    \label{fig:summary}

    \begin{abstract}
      \textit{Vision transformers (ViTs) - especially feature foundation models like DINOv2 \rebuttal{and DINOv3} - learn rich representations useful for many downstream tasks. However, architectural choices (such as positional encoding) can lead to these models displaying positional biases and artefacts independent of semantic content. This makes zero-shot adaption difficult in fields like material science, where images are often cross-sections of homogeneous microstructure (i.e. having no preferred direction). In this work, we investigate the positional bias in ViTs via linear probing, finding it present across a range of objectives and positional encodings, and subsequently reduce it by finetuning models to use ALiBi relative positional encoding. We demonstrate that these models retain desirable general semantics and their \rebuttal{less positionally biased} features  can be used successfully in trainable segmentation of complex microscopy images.}
    \end{abstract}
\end{strip}

\let\thefootnote\relax\footnotetext{\textsuperscript{*}samuel.cooper@imperial.ac.uk}

\lhead{\scshape Pawlowsky \textit{et al.}}
\chead{\scshape What DINO saw}
\rhead{\scshape Preprint}




\newpage

\section{Introduction}
\label{sec:intro}


Vision transformers (ViTs) have become ubiquitous in recent years, with state-of-the-art performance across benchmarks \cite{COCA_VIT}, easy multi-modal integration \cite{CLIP} and strong scalability \cite{SCALING_VIT}.
Key to this success are the feature foundation models: large backbones, trained via self-supervised learning (SSL) on massive corpuses of images, whose frozen features can be used to train small head networks or in pipelines with no further training, such as $k$-means clustering.
Examples include the masked autoencoder (MAE), iBOT and the DINO \& JEPA series of models \cite{MAE,iBOT,DINO,DINOv2,DINOv3,I_JEPA}; each has different self-supervised objectives and subsequent representations \cite{MAE_VS_DINO_SSL}, but all display richer and more robust representations than comparable supervised models \cite{DENSE_VIT}.

The success of feature learning extends across domains - various foundation models exist for (bio)medical imaging, satellite data or robot manipulation - as well as across tasks, existing as backbones inside task-specific foundations models such as SAM \cite{SAM} and EVA-CLIP \cite{EVA_CLIP}.
Most of these models (regardless of training objective) use the standard ViT architecture: a patch projection, positional encoding (PE) and some number of attention layers.

Architectural choices clearly influence output feature quality: background attention sinks necessitated the addition of register tokens \cite{VIT_REGISTERS}, whilst others have suggested the learned positional encoding is responsible for pernicious `feature artefacts' and trained denoisers to remove them \cite{DVT}.
Contemporaneous work has analysed the flow of information through these models, specifically in terms of the preservation and interaction of positional information with semantic information \cite{RABBIT_HULL,BFD}.
The quality and consistency of these ViT features is important for pipelines which use them without additional training, like LOST \& MOST's unsupervised object detection \cite{LOST,MOST,UNSUP_OD_SURVEY}, various (spectral) clustering methods for semantic segmentation \cite{DEEP_SPECTRAL,TOKENCUT, OD_NORMCUT}, and user-guided `trainable segmentation' of images \cite{FEATURE_FOREST, CONVPAINT, HR_DV2, VULTURE}.

The application of computer vision to materials science imaging suffers from several challenges: scarcity of open data, a wide range of length scales and materials systems, and the ambiguities of electron microscopy \cite{MAT_SCI_DATA_CHALLENGE}.
This makes the general, `off-the-shelf' features of foundation models appealing (especially in low- or no-training pipelines), and several works have leveraged them successfully \cite{MAT_SCI_FEATURE_VIT, HR_DV2, VULTURE, FEATURE_FOREST, CONVPAINT, dino_feats_mitochondria_seg, vfm_micro_review, nanotube_quant}.
However, materials science images, typically captured with a Scanning or Transmission Electron Microscope (SEM/TEM), are fundamentally different to natural images: they are grayscale, are sometimes hundreds of megapixels large, and are often homogeneous cross-sections. 
This homogeneity is especially challenging given the clear positional biases seen in ViTs like DINOv2, which has been show to have negative impacts on trainable segmentation \cite{HR_DV2}.  

In this work, we seek to reconcile these differences by finetuning a ViT (\textit{i.e.} DINOv2) with 2D-aware ALiBi positional encoding \cite{ALIBI}. 
First, we characterise the positional bias across various literature ViTs using linear probes, finding some channels of the output features which are almost purely positional ramp functions regardless of input image. These exist in both MAE and DINO models, and across different positional encodings (learned or RoPE), though not in supervised ViTs.

Next, we finetune ViT checkpoints (DINOv2 \& DINOv3) to use 2D-aware ALiBi positional encodings, with cylindrical boundary conditions and normalisation to enable positional encoding interpolation \cite{ALBI_POS_INTERP}.
Surprisingly, once the learned PE is replaced with the \rebuttal{less biased}  ALiBi PE it is sufficient to use the (biased) embeddings of the original ViT as a training target to recover semantics. 

Finally, we benchmark the performance of these finetuned variants in terms of both their homogeneity and linear-probe performance on standard segmentation benchmarks (VOC, ADE20K), finding that performance is largely maintained and even improved in some cases. We further apply our ALiBi models to weakly- or unsupervised tasks ($k$-means clustering and trainable segmentation) and find that their homogeneous features improve the resulting segmentations.

\section{Background}
\label{sec:background}

\subsection{Vision Transformers}
Transformers are artificial neural networks which process a series of `tokens' (projections of discretisations of some input sequence $T = \{\textbf{t}_1, \textbf{t}_2, ..., \textbf{t}_N\}$) which learn to update the representation of some $d$ dimensional token $\textbf{t}_i$ with respect to all other tokens via attention,

\begin{equation}
    \text{Attn}(\textbf{Q},\textbf{K},\textbf{V}) = \text{softmax}(\frac{\textbf{Q}\textbf{K}^\intercal}{\sqrt{d}})\textbf{V}
\end{equation}

where $\textbf{Q}, \textbf{K}, \textbf{V}$ are packed matrices of the query, key and value learned projections of $T$ \cite{ATTN, ALGO_TRANSFORMERS}.

Vision transformers discretize an input image into $s \times s$ px non-overlapping patches and perform a learned linear projection to map from $\mathbb{R}^{B \times C \times H \times W} \rightarrow \mathbb{R}^{B \times d \times N_t}$ where $N_t = \text{number of tokens} = HW/s^2$ \cite{ViT}.
This is done to avoid the overhead of processing $HW$ pixels via the $n^2$ attention mechanism. 

The patch projection followed by multiple attention layers is common across most ViTs. 
ViTs tend to have a `size': (S)mall, (B)ase, (L)arge \textit{etc.}, which have corresponding hidden dimension $d$ of 384, 768, 1024 \textit{etc.}, as well as having a greater MLP dimension and more attention heads.

\subsection{Positional encodings}
Attention is fundamentally unordered, so transformer models attach a `positional encoding' in order to associate each token with its position in a sequence \cite{ATTN}. Typically these are injected (\textit{i.e.} added to the patch's linear projection) at the start of the model, and the model then learns to use them alongside token content across the layers of the network.

Early work with vision transformers experimented with simple positional encodings, such as raster-scan order (an increasing integer index from left-to-right, top-to-bottom), sinusoidal (adding sinusoids of increasing wavelength along the channel dimension) and learnable 1D encodings \cite{ViT, SIN_POS_ENC}. These are absolute positional encodings: they uniquely identify each token with a given position in the original image. Note that sinusoidal encodings can theoretically uniquely identify tokens in a sequence up to the largest wavelength added. 

Different encodings, such as 2D-learned, relative positional encodings or no positional encoding (NoPE), were tried, but did not improve benchmark performance and so were not used \cite{ViT}.
To generalise to new image sizes, the positional encoding can be interpolated with respect to the new sequence length.

\subsection{Relative positional encodings and ALiBi}
Relative positional encodings (RPE) encode position information in terms of relative distances between two tokens. 
This can be additional relative attention, where the query content attends to relative positional keys \cite{ViT}, or in the recently popular Rotary Positional Encodings (RoPE) \cite{RoPE} a rotation of query-key dimension pairs proportional to token positions, such that only relative position are expressed inside the attention inner product.
Similar to sinusoidal encodings, the rotation applied to dimension pair $d$ of a token $t$ decreases as $d$ increases.

Attention with Linear Biases (ALiBi) introduced positional encodings as linear offsets added to token-token attention scores as a function of their relative distance \cite{ALIBI}.
Concretely, the ALiBi attention scores for the $i$th query $\textbf{q}_{i}$ are
\begin{equation}
    \text{softmax}(\textbf{q}_{i}\textbf{K}^{\intercal} + \textit{m} \cdot [-(i-1), ..., -2, -1, 0]),
\end{equation}
where $m$ is a head-specific scalar, which can be learned or set according to a fixed geometric progression \cite{ALIBI}.
This encourages different heads to focus on different distances. ALiBi enforces `an inductive bias towards recency' (for images, proximity) and operates directly on attention scores, rather than encoding information in a token's hidden state \cite{ALIBI}.
ALiBi positional encodings can also benefit from the `interpolation trick' in order to improve length generalization \cite{ALBI_POS_INTERP}.

A few ViTs have integrated relative positional encodings: MViTv2 uses decomposed RPEs \cite{MViTv2}, DINOv3 uses a modified RoPE variant \cite{DINOv3} and domain-specific models like CROMA (in satellite imaging) \cite{CROMA} and Long-MIL (for histopathology imaging) \cite{HISTO_ALIBI} have used 2D-aware ALiBi PEs.
This choice of RPE improved various properties such as shift-equivariance and length generalization. 

\subsection{DINO \& feature learning}
The scalability of transformers and of SSL led to the development of large, general-purpose `feature foundation models'. 
One of the earliest examples is DINO (self distillation with no labels), which showed that self-supervised vision transformers learnt meaningful decompositions of images in a way that similar convolutional neural networks did not \cite{DINO}.

DINOv2 repeated the approach on a large curated dataset, producing `all-purpose visual features' that enabled state-of-the-art performance on benchmarks with linear probes \cite{DINOv2}.
DINOv3 applied various tricks (such as Gram anchoring) to produce dense, coherent feature maps for high-resolution images \cite{DINOv3}.
Both DINO and DINOv2 used learned 1D positional encodings, whereas DINOv3 used RoPE positional encoding, ostensibly to avoid positional leakage into the features \cite{DINOv3, DV3_GH_POS_ENC}.

\subsection{Feature artefacts \& positional bias in ViTs}
Despite their impressive performance on benchmarks, the DINO features (and ViT features more broadly) exhibit certain biases and artefacts. 
The authors of ref. \cite{VIT_REGISTERS} found anomalous high-norm background tokens caused by the model's re-use of them for computation. 
They ameliorated this by adding scratchpad `register tokens' ($\texttt{[REG]}$) which were discarded at the end of the forward pass. 

`Denoising vision transformers' (DVT) \cite{DVT} noted that feature artefacts were largely independent of semantic content and were correlated with the featuremap of the zero-input tensor, and suggested a link between the positional encoding and these artefacts as a possible explanation. 
Weak artefacts were noted even in a DINOv2 checkpoint with $\texttt{[REG]}$ tokens. 
They trained an additional denoiser network to remove these artefacts. 
We noted these denoised feature map PCAs also appeared to display less left-to-right positional gradients, and therefore included them as a comparison in \cref{sec:results}.

Recent work has examined the role of positional information in feature representation in ViTs. 
One study demonstrated that tokens were locally connected even after projection into a lower-dimensional non-positional basis as evidence for a `linear representation hypothesis' of ViT features \cite{RABBIT_HULL}.
Another study examined the role of position and content interactions inside ViTs, pointing to `low-energy' positional modes inside the features which diminish in importance through subsequent layers of the network \cite{BFD}.
They further showed how the positional structure of DINOv2 was preserved as a 2D sheet across layers, whereas supervised ViTs collapsed this information into a `quasi-1D structure'.

\subsection{Trainable segmentation} 
Trainable segmentation (also called `interactive segmentation' or `pixel classification') is a weakly supervised segmentation technique that trains a classifier to map from a set of features describing pixels to user-drawn labels \cite{WEKA, ILASTIK}.
Popular implementations use a classical multi-scale feature set (Gaussian blurs, Sobel or Laplacian edge detection, Hessian filters \textit{etc.}) and a lightweight fast classifier like a random forest (capable of modelling non-linear interactions) \cite{WEKA, ILASTIK}.
Once trained, the classifier can be applied to unlabelled pixels to produce a segmentation.

Trainable segmentation tools are often used in bio-image segmentation \cite{ILASTIK, WEKA, ILASTIK_CELL_COUNTING, ILASTIK_HISTO}, and occasionally applied in materials science imaging \cite{WEKA_STEELS, WEKA_SHALE, KINTSUGI}.
One drawback of the approach was the relative inexpressiveness of the classical features when applied to more complex features \cite{HR_DV2}, such as crack networks, imaging artefacts (beam damage, charging, curtaining in electron microscopy data) or the `pore-back' effect \cite{KINTSUGI} in porous media.

Several works aimed to fix this problem by combining the classical and `deep' features of ViTs and CNNs in a trainable segmentation workflow, which produced better segmentations of complex microscopy images \cite{FEATURE_FOREST, CONVPAINT, HR_DV2, VULTURE}.
However, some noted the difficulties presented by the positional biases of ViTs, especially in the materials science context where researchers frequently image large cross-sections of homogenous materials \cite{HR_DV2}.
Attempts to reduce these positional biases by averaging over transformations (flips, rotations) were moderately successful but failed to reduce centrosymmetric biases and came with associated runtime costs \cite{HR_DV2}.

On the nano- and micro- length scale, these materials are effectively semi-infinite random arrangements of microstructural features, often with no central focus or preferred direction - a classifier fit to positional information inside the features is therefore often wrong.
There is, in essence, a mismatch between the architectural biases and training data of the models, and the images researchers wish to analyse.

\section{Method}
\label{sec:method} 

\subsection{Positional linear probing} 
\label{sec:method:lin_probe}

\begin{figure*}
\centering
    \includegraphics{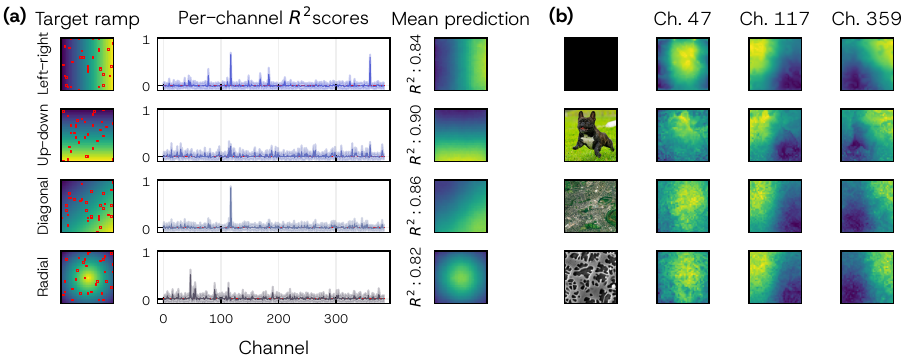}
    \caption{Linear probe analysis of DINOv2-S features. \textbf{(a)} We train linear probes to map from image features (or individual channels) to randomly sampled (red squares) ramp functions, reporting $R^2$ scores on holdout regions. Per-channel scores and predictions (which use all channels) are both averaged over a dataset of \rebuttal{100} homogenous microscopy images, \textit{i.e.} images with no preferred direction. \textbf{(b)} The channels with the highest positional $R^2$ scores for a series of images, including a satellite image and an electron microscope image of a nickel superalloy \cite{NI_SUPERALLOY}.}
    \label{fig:linear_probe}
\end{figure*}

One simple way to analyse positional biases in ViTs is via linear probing of output features. 
If positional information is linearly present in combination (or worse, in isolation) in the output feature channels, then any secondary classifier (linear/logistic regressor, random forest, XGBoost, \textit{etc.}) can fit to it.

We trained linear probes to map from both the full feature stack and single channels of output and layer-wise ViT features to a series of 1D ramp functions: left-right, up-down, diagonal and radial. 
Ramp values were normalised from 0-1 across the image.
Probes were fit over a randomly sampled subset of spatial patches, and $R^2$ was measured over holdout regions \rebuttal{(i.e. over the patches not used during training)}. 
In practice, around $2.5\%$ of patches were sampled to train the probe.
The ramps and example samplings (\rebuttal{red squares}) can be seen in \cref{fig:linear_probe}.

\rebuttal{Values were averaged over a dataset of 100 microscopy images (micrographs), specifically random samples drawn from a 22,000 SEM image dataset} \cite{BIG_SEM_DATASET}.
\rebuttal{These images were resized to a longest-side length of 1024 px, and then centre-cropped.}
\rebuttal{We present some examples in \cref{supp:sec:linear_probe:dataset}}.
We tested the same linear probing setup on two additional datasets: \rebuttal{100} random textures (of everyday objects) from the `Describable Textures Dataset' (DTD) \cite{DTD} and \rebuttal{100} instances of uniform white noise. \rebuttal{6} repeat measurements (with different samplings) were made for each image.

Spatialised subsampling was tested (\textit{i.e.} with holdout vertical and horizontal regions) but did not make a significant impact, neither did choice of regularised (Ridge) or unregularised regressor.
The same trends were observed when using a non-linear classifier (shallow MLP) as with a linear model.

To produce a single numerical positional bias `score', multivariate linear regression was performed to jointly predict the normalised $(x,y)$ coordinates from the full feature stack.

\subsection{ALiBi finetuning}
\label{sec:method:alibi_dv2}

Being injected at the very start of the network means positional encodings are integral to the learned representations of ViTs. We experimented with schemes to adapt the existing learned positional encoding to make it more homogeneous (tiling, zooming, \textit{etc.}) but were unsuccessful: either output features were degraded or still displayed edge effects. 
This motivated some form of further training with a positional encoding that was homogeneous by construction.

We found RoPE models could still express positional bias (see \cref{sec:results:linear_probe} and \cref{sec:results:toy_model}), and therefore elected to use ALiBi encodings. 
We used Euclidean distance as the distance metric with wrapped boundary conditions in order to avoid asymmetries in the distance matrix. 
We used a fixed $m=1$ for all head-specific scalars to avoid potential asymmetries compounding across layers (we examine the impact of trainable heads in \cref{sec:supp:model:trainable_m}). 
ALiBi offsets were added to each attention layer.
The entries in the ALiBi distance matrix were normalised between 0-1 by dividing by the longest distance.
The learned PE was frozen and set to zero during finetuning.

In terms of training target, we found it was sufficient to use the embeddings of the original (biased) ViT (DINOv2-S or DINOv3-S+) as a teacher. 
Our motivation was as follows: those embeddings had many of the desirable general semantics we wished to preserve, they were simple and cheap to generate, and the setup of our ALiBi PE made it difficult for our model to express those biases, even when present in the data.
We experimented with producing `cleaned' target embeddings by directly averaging over the features of transformations of the input images, but these were costly to produce and did not improve performance.
During training, we zeroed the four most positional channels as identified in \cref{sec:results:linear_probe} - this reduced the re-emergence of weak long-range biases.

The dataset used to generate embeddings was COCO-Stuff \cite{COCO_STUFF}.
Images were sampled with shortest-side resizing to 224 px followed by a center crop to (224, 224) px. 
No training augmentations were used, though these could be added in future to improve properties, such as rotation equivariance.
Following DINOv2, we performed a short period of multiscale training at (518, 518) px resolution \cite{DINOv2}.
We found this was important to improve length generalisation of the model - see \cref{sec:supp:model:multiscale_training} for more details.

Further training details (learning rates, batch size, \textit{etc.}) are available in \cref{sec:supp:model:hyperparams}. As a baseline comparison for the benefits of ALiBi encoding, we also finetuned a DINOv2 checkpoint with no positional encoding (\textit{i.e.} a bag of patches). \rebuttal{To do this, we zero'd and froze the model's original PE, and did not replace it.}
We refer to this model as `NoPE' in the following sections. The training setup for ALiBi-Dv2 and NoPE were identical.

To demonstrate our approach generalises, we extended this training regime to DINOv3-S+, again blanking the four most positional channels as identified by the linear probing, all other training parameters were kept the same; we call this model `ALiBi-Dv3'.

\section{Results and discussion}

In this section we characterise the positional biases across literature ViTs via linear probing, and how this motivates the training of a homogeneous feature model (ALiBi-Dv2/ALiBi-Dv3). We proceed to examine the properties of our ALiBi-finetuned models in terms of feature PCAs, linear probe semantic segmentation (VOC, ADE20K) and weakly-supervised segmentation of materials microscopy images.

Unless otherwise stated, we used DINOv2-ViT-S (patch size 14) with \texttt{[REG]} tokens as `DINOv2', DINOv3-ViT-S+ (patch size 16) as `DINOv3' and DINOv2-ViT-S + \texttt{[REG]} +  Denoising Vision Transformer's general lightweight denoiser layer as `DVT'.

\label{sec:results} 
\subsection{Positional linear probing}
\label{sec:results:linear_probe}

\begin{figure}
\centering
    \includegraphics{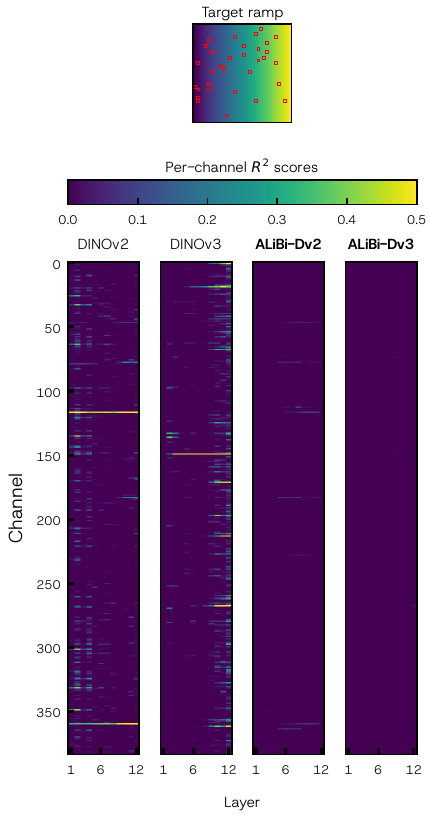}
    \caption{Per-channel per-layer `positional fingerprint' of $R^2$ scores for DINOv2, DINOv3, ALiBi-Dv2, \rebuttal{and ALiBi-Dv3} for a left-right target ramp. DINOv2 begins with positional information spread across channels (its learned PE is added at the start of the network), which later decreases, whereas for DINOv3 the channels become more positional with layer depth (RoPE is applied at each layer). \rebuttal{Our ALiBi-models have less positional information present across their channels and layers.}  }
    \label{fig:positional_fingerprint}
\end{figure}

\null\par

\begin{table*}
    \centering
    \begin{tabular}{l c c c c c c}
    \toprule
        Model & \textit{d} & PE & Supervised & $R^2_{\text{micro}}$ & $R^2_{\text{texture}}$ & $R^2_{\text{noise}}$ \\
    \midrule
        \multirow{2}{*}{DINO} & S & Learned & $\times$ & \cellcolor{red!25} 0.66 & \cellcolor{red!25} 0.51 & \cellcolor{red!25} 0.52 \\
             & B & Learned & $\times$ & \cellcolor{red!25} 0.71 & \cellcolor{red!25} 0.57 & \cellcolor{red!25} 0.52 \\
        \multirow{2}{*}{DINOv2} & S & Learned & $\times$ & \cellcolor{red!25} 0.86 & \cellcolor{red!25} 0.70 & \cellcolor{red!25} 0.89 \\
             & B & Learned & $\times$ & \cellcolor{red!25} 0.76 & \cellcolor{red!25} 0.57 & \cellcolor{red!25} 0.89 \\
        DINOv3 & S+ & RoPE & $\times$ & \cellcolor{red!25} 0.97 & \cellcolor{red!25} 0.90 & \cellcolor{red!25} 0.98 \\
        ViT (MAE) & B & Learned & $\times$ & \cellcolor{red!25} 0.79 & \cellcolor{red!25} 0.60 & \cellcolor{red!25} 0.70 \\
        SAM & 256 & Learned & $\checkmark$ & \cellcolor{red!25} 0.77 & \cellcolor{red!25} 0.64 & \cellcolor{red!25} 0.91 \\
        ViT (IN1k) & B & Learned & $\checkmark$ & \cellcolor{green!25} 0.07 & \cellcolor{green!25} -0.01 & \cellcolor{green!25} -0.05 \\
        DEiT & B & Learned & $\checkmark$ & \cellcolor{green!25} 0.21 & \cellcolor{green!25} 0.10 & \cellcolor{green!25} 0.01 \\
        CLIP & B & Learned & $\checkmark$ & \cellcolor{green!25} 0.12 & \cellcolor{green!25} 0.03 & \cellcolor{green!25} -0.06 \\
        EVA02 & B & RoPE & $\checkmark$ & \cellcolor{green!25} 0.09 & \cellcolor{green!25} -0.23 & \cellcolor{green!25} -0.10 \\
    \bottomrule
    \end{tabular}
    \caption{\rebuttal{$(x, y)$ linear probe $R^2$ scores for a set of existing models which produce patch-level features. Supervised models - including those with an SSL backbone - have lower $R^2$ scores. Scores are reported per-dataset: `micro' for micrographs, `texture' for DTD \cite{DTD} texture images and `noise' for random noise.}}
    \label{tab:linear_probe_literature}
\end{table*}

\begin{table*}
    \centering
    \begin{tabular}{l c c c c c}
    \toprule
        Model & \textit{d} & PE & $R^2_{\text{micro}}$ & $R^2_{\text{texture}}$ & $R^2_{\text{noise}}$ \\
    \midrule
        \multirow{2}{*}{DINOv2} & S & Learned & \cellcolor{red!25} 0.86 & \cellcolor{red!25} 0.71 & \cellcolor{red!25} 0.89 \\
             & B & Learned & \cellcolor{red!25} 0.76 & \cellcolor{red!25} 0.56 & \cellcolor{red!25} 0.89 \\
        DINOv2 (CB) & S & Learned & \cellcolor{red!25} 0.84 & \cellcolor{red!25} 0.67 & \cellcolor{red!25} 0.88 \\
        DVT & S & Learned & \cellcolor{red!25} 0.80 & \cellcolor{red!25} 0.68 & \cellcolor{red!25} 0.83 \\
        \textbf{ALiBi-Dv2} & \textbf{S} & \textbf{ALiBI} & \cellcolor{green!25} \textbf{-0.06} & \cellcolor{green!25} \textbf{0.26} & \cellcolor{green!25} \textbf{-0.91} \\
        \textbf{ALiBi-Dv3} & \textbf{S+} & \textbf{ALiBI} & \cellcolor{green!25} \textbf{-0.31} & \cellcolor{green!25} \textbf{0.17} & \cellcolor{green!25} \textbf{-0.94} \\
    \bottomrule
    \end{tabular}
    \caption{\rebuttal{$(x, y)$ linear probe $R^2$ scores for DINOv2 variations. `CB' refers to zeroing the 4 most positional channels of DINOv2 (see \cref{fig:linear_probe}). Our ALiBi model has more homogeneous features and therefore has a lower $R^2$.}}
    \label{tab:linear_probe_ours}
\end{table*}

Given that positional biases had been observed in DINOv2, we started by performing linear probing on its output features for a series of ramp functions, plotting the result in \cref{fig:linear_probe}. The results were averaged over a dataset of theoretically homogeneous microscopy images: this was to avoid true horizontal or vertical gradients (\textit{i.e.} sunsets, the sky, field-of-view) skewing the results. 

Of note are the distinct peaks in the per-channel $R^2$ scores, indicating that some channels were highly correlated with the different 1D ramps. 
We display these specific channels across a set of different images and see that the output features are consistently positional (radial or diagonal gradients) regardless of image content. 
When predicting using all channels of the feature stack, the resulting $R^2$ is high, \textit{i.e.} consistently 0.8 or above.

To identify how widespread the problem was, we then measured the positional bias of a series of literature ViT models via linear probing, presenting the results in \cref{tab:linear_probe_literature}. 
As noted in \cref{sec:method:lin_probe}, we regressed onto the $(x,y)$ 2D ramp function to produce a single $R^2$ value that captured both left-to-right and up-to-down biases.

We found biases across all models in the DINO series, including DINOv3, which used a relative positional encoding (RoPE).
We also observed biases in a ViT trained with masked autoencoding \cite{MAE}.
Other supervised models like DEiT \cite{DEIT}, CLIP \cite{CLIP} and EVA02 \cite{EVA02} were found to have much lower positional bias, and absence of problematic positional channels.
This supports Ref. \cite{BFD}'s claim that self-supervised models preserve positional information more strongly than supervised models.

The relatively high positional score of the SAM backbone complicates the picture: the backbone was initially pretrained with SSL as an MAE ViT, but was further trained in a supervised manner alongside the decoder head to perform promptable segmentation.
It is unclear if this high positionality is a holdover from the SSL pretraining, or the fact that it was supervised to perform a dense task (segmentation) compared to the other supervised models, which performed more global tasks (classification or captioning).

When repeating the probes on the DTD we found the $R^2$ scores to be around 0.05-0.1 points lower for most models, though the trends remained the same (including the same outlier channels). 
The random noise ramps had a similar set of $R^2$ scores to the micrograph dataset - the biases appear to be stronger (or less mixed with genuine semantics) in out-of-distribution data.

DINOv3's high positional score was surprising, given the series of adjustments made to the positional encoding (axial RoPE, positional jitter) to improve length generalization and reduce positional leakage \cite{DINOv3, DV3_GH_POS_ENC}.
To try and understand why this was the case, we performed a left-to-right linear probe per-channel at each layer of DINOv2 and DINOv3 (again averaged across 15 images), and plotted the results in \cref{fig:positional_fingerprint}. 

\begin{figure*}
\centering
    \includegraphics{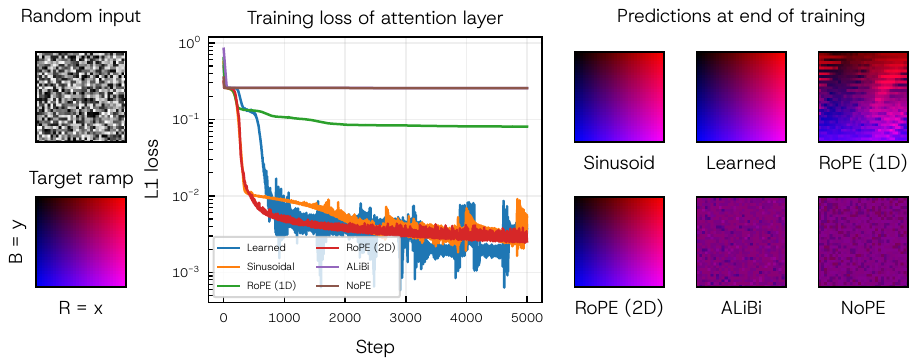}
    \caption{Training of a toy attention layer for a series of positional encodings: fixed sinuoidal, learned, RoPE (1D and 2D), ALiBi and NoPE. The goal is to test whether the layer can learn a mapping from a fixed random input to an $(x,y)$ grid using its positional encoding and weights. Absolute encodings (sinusoidal, learned) can, as can both RoPE variants. ALiBi and NoPE, by contrast, cannot - this makes them good candidates for post-hoc finetuning.  }
    \label{fig:toy_model}
\end{figure*}

We found DINOv2 began with positional correlations spread across multiple channels, which diminished in the later layers of the network, whereas DINOv3's positional correlations increased throughout the layers. 
This can be explained by the structure of their PEs: DINOv2's learned PEs are injected once at the start, and DINOv3's RoPE is applied at each layer. Both models had a small number of highly positional outlier channels.

That DINOv2 loses some positional information throughout its layers has been noted previously by Ref. \cite{BFD}, who trained linear probes to predict exact token position and found the accuracy decreased throughout the network.
Whilst our findings in \cref{fig:positional_fingerprint} do support this, we also find that the DINOv2's $R^2$ when using all channels remains relatively consistent across layers: the model is able to express these biases even if it loses token-perfect information.
We present additional layer-channel plots in \cref{supp:fig:fingerprints_diff_models}.
Further works examining how these positional biases arise during training, especially across different objectives and positional encodings, is an interesting avenue of future research. 

Next, we compared approaches for reducing the positional bias of DINOv2 and the results are presented in \cref{tab:linear_probe_ours}. The approaches were `channel-blanking' (CB), where the four most positional channels of DINOv2 were set to 0 on forward pass, and DVT, the learned denoiser layer. Both reduced (but did not eliminate) the positional bias. We found DVT reduced left-to-right biases but emphasised up-to-down biases. 

The failure of these approaches to eliminate the positional bias motivated the training of ALiBi-DINO variants as described in \cref{sec:method:alibi_dv2}. We report the linear probe results for our ALiBi models in \cref{fig:positional_fingerprint} and \cref{supp:fig:alibi_probe}, as well as \cref{tab:linear_probe_ours}. We find greatly reduced positional bias scores across all channels and layers of the network. Additional positional probe plots for different ViTs are available in \cref{supp:sec:linear_probe:more_models}.

Note that a negative $R^2$ arises from applying a model fit on a subset of the features (red squares) to the other features, which may be drawn from a different distribution. It does not represent anti-correlation. 

\subsection{Which encodings admit biases?}
\label{sec:results:toy_model}

To explain the results of \cref{sec:results:linear_probe}, we implemented a toy model - a single layer with an input projection, $q,k,v$ projections plus attention calculation, an output MLP projection, and a set of different positional encodings (learned, fixed 2D sinusoidal, 1D and 2D RoPE, ALiBi, and NoPE).

The input was a $32 \times 32$ fixed random tensor, and the target was a normalised $(x,y)$ ramp.
The model had a hidden dimension of $d=32$, 4 attention heads, and was trained with a learning rate of $5 \times 10^{-4}$ with MAE loss for 5000 epochs.
Both the sinusoidal and RoPE schemes had the standard base wavelength of 10,0000.
If the toy layer with a given positional encoding can learn to reproduce the ramp function with this homogenous random input, then we assume it is able to learn and express them more generally, especially as part of a larger model.

We present the results of this in \cref{fig:toy_model}. As expected, the absolute positional encodings (learned, sinusoidal) can trivially learn the mapping. Notably, both variants of RoPE can learn to reproduce the grid - this suggests that, despite being a relative positional encoding, it is capable of modelling long-range positional biases, potentially in low-frequency rotations.

ALiBi and NoPE are not able to model the positional biases - this supports our findings in \cref{tab:linear_probe_literature} and \cref{tab:linear_probe_ours}, and motivates our choice of positional encoding for the finetuning. That NoPE cannot learn the biases suggest that the toy model is not capable of memorization \textit{i.e,} learning a lookup from pixel values to ramp values).

 \subsection{Qualitative feature PCAs}

\begin{figure*}
\centering
    \includegraphics{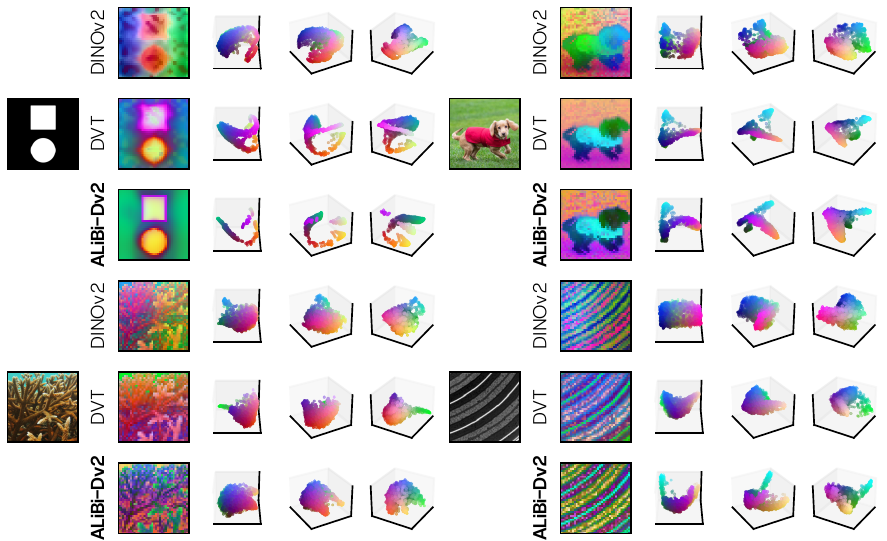}
    \caption{Feature PCA comparisons for DINOv2-S, DVT and our ALiBi-Dv2. ALiBi-Dv2 produces features which are semantically rich but that display less positional bias and artefacts. The token geometry is (locally) smooth, and in many cases similar to that of DVT. Of note is the `square-circle' image, which retains `objectness' across indistinguishable interior patches, the preserved vertical bias of the dog image (from depth-of-field) and the reduced positional biases on the X-ray CT cross-section of a pintype Li-ion battery \cite{BIL}.}
    \label{fig:3d_pcas}
\end{figure*}

\cref{fig:3d_pcas} shows feature PCA visualisations for a series of images. 
Visualisations were produced by projecting output patch features from $d=384 \rightarrow d'=3$ and plotting the 3 components as RGB channels as per DINOv2 \cite{DINOv2}.
PCAs are fit per-image; images are resized to $(518, 518)$ resolution.

From these PCAs we can observe reduced positional gradients (both left-right and up-down) for ALiBi-Dv2 compared to DINOv2 and DVT.
For the synthetic `square-circle' example the resulting features are sharper and more separated, whereas for DINOv2 and DVT, albeit to a lesser extent, the square and circle features appear closer, either due to the influence of the positional encoding or weak artefacts. 
For some images, such as the dog, the PCAs of DVT and ALiBi-Dv2 are similar; this was more often seen in in-distribution images, where the positional bias was expressed less strongly (in relative terms) to the semantic content.

The `token geometry' (3D plot of patch features according using PCA coordinates) remains locally smooth echoing the results of Ref. \cite{RABBIT_HULL} when projecting DINOv2 features into a learned apositional subspace.

\begin{figure*}
\centering
    \includegraphics{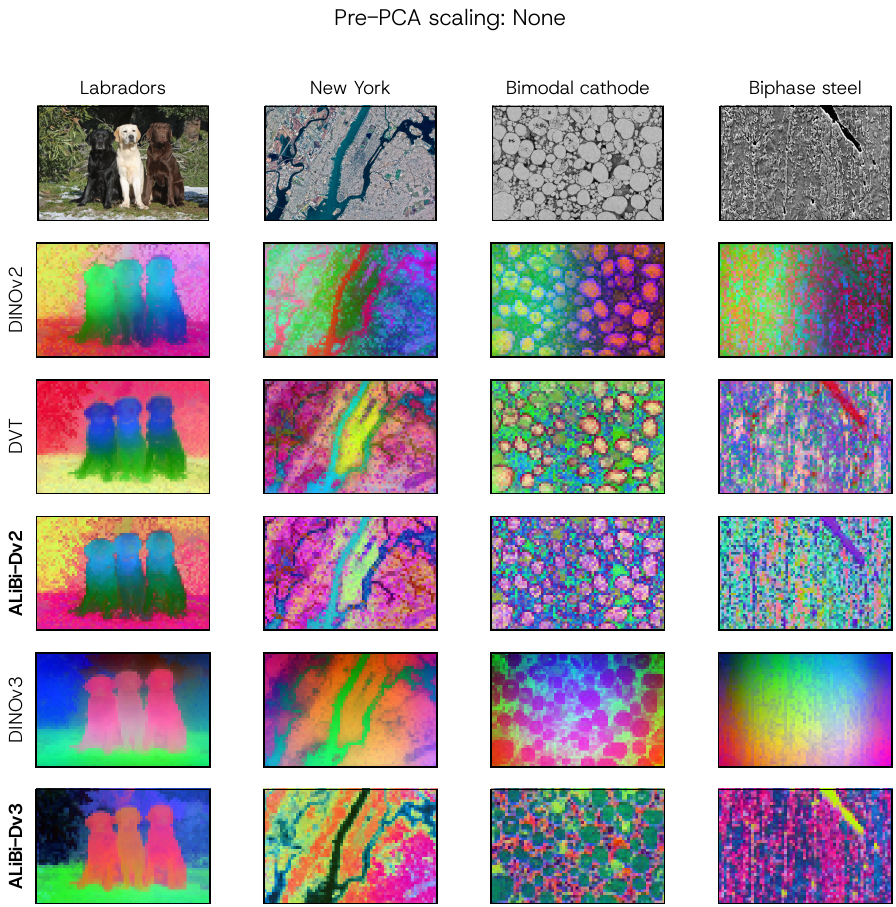}
    \caption{Feature PCA visualisations for DINOv2, DINOv3 and our ALiBi-Dv2 \& ALiBi-Dv3. We retain desirable object decompositions (e.g. head vs body of the dogs), whilst having less positional features for the satellite image and the SEM images of a battery cathode \cite{BIL} \& biphase steel \cite{BIPHASE_STEEL}.}
    \label{fig:more_pcas}
\end{figure*}

\cref{fig:more_pcas} shows further PCA visualisations on larger rectangular images, this time for DINOv2, DINOv3 and our ALiBi models. Again we see positional biases displayed in the features (some edge effects for DINOv2, ramps for DINOv3), which become more problematic the less `natural' the images become (\textit{i.e.} dogs vs satellite photography vs electron microscopy). Agreeing with the results of \cref{sec:results:linear_probe}, we see that DINOv3 appears more positional than DINOv2.
Contemporaneous work has also supported our observation that DINOv3 has stronger positional biases than DINOv2 \cite{homog:insid3}.

Once again, we note that the ALiBi models retain the ability to decompose large central objects, like the dogs' heads and bodies, whilst retaining a high degree of homogeneity on more homogeneous images. The `bimodal cathode' is a scanning electron microscope image of a lithium-ion battery cathode with a mix of different particle sizes - these have better packing and therefore promise higher energy densities. Analysing them is therefore of great interest, and the ALiBi models produce homogenous features of complicated microstructural concepts such as particle size, cracking and gaps (pores) in the structure.   

We note that PCA visualisations are imperfect tools: a feature having less variance and therefore not being present in the first 3 components does not imply the feature is absent, and preprocessing choices can affect what features are shown. However, this does become a practical problem if the PCA is used as part of an unsupervised workflow, such as segmentation or $k$-means, or if trying to filter out the positional modes based on variance or energy. See \cref{supp:fig:pca:more_pca_std} for feature PCAs with scaling.

It is reasonable to ask the following question: does the improved homogeneity after finetuning stem from the ALiBi encoding specifically, or is it that altering the positional encoding is sufficiently damaging to the model's representations that it cannot re-learn the biases during finetuning, regardless of choice of encoding? As a simple control, we repeat our finetuning approach as described in \cref{sec:method} (using DINOv2 with channel blanking and same hyperparameters), but instead of adding ALiBi encoding, we add a fixed sinusoidal positional encoding.

As expected, the finetuned sinusoidal checkpoint is capable of re-learning the biases present in the original features. We present feature PCAs comparisons between DINOv2, a sinuosidal DINOv2, a NoPE DINOv2 and our ALiBi-Dv2 in \cref{supp:fig:nope_v_alibi}.
That the channel-blanked sinusoidal model recovered biases indicates channel blanking alone is not sufficient to eliminate positional biases during finetuning, it must be coupled with a \rebuttal{less biased} positional encoding (\textit{i.e.} ALiBi).

\subsection{Feature properties: robustness, similarity, decomposition}

\begin{figure*}
\centering
    \includegraphics{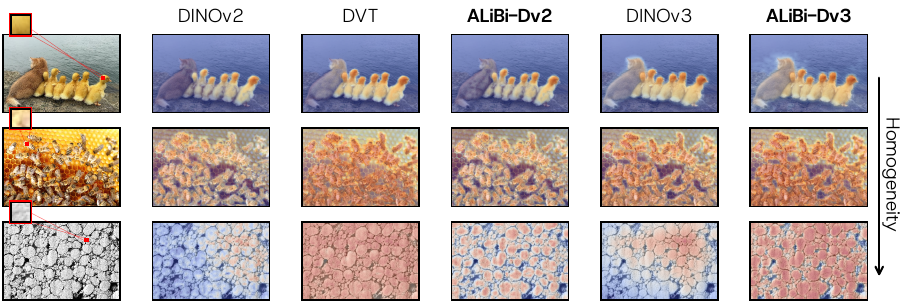}
    \caption{Cosine similarity scores for DINOv2 and our ALiBi models for a given query patch (red inset square). ALiBi-Dv2 and -Dv3 have more uniform similarities for increasingly homogenous (and out-of-distribution) images, such as the SEM image of a Li-ion battery cathode \cite{BIL}.}
    \label{fig:cosine}
\end{figure*}

\begin{figure}
\centering
    \includegraphics{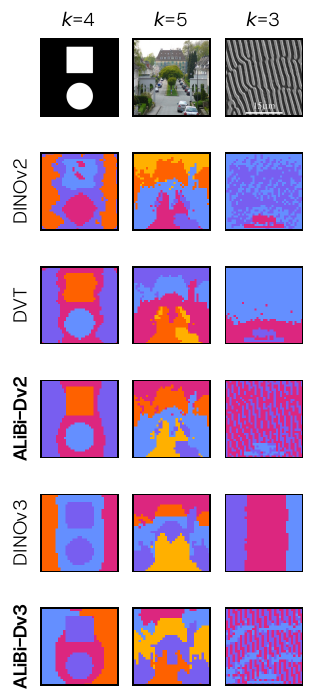}
    \caption{{k}-means clustering of features of DINOv2-S, DVT, DINOv3, and our ALiBi models. For some images, like the SEM micrograph of a eutectic alloy \cite{DOITPOMS}, both DINOv2 and DVT decompose using positional information (radial or vertical), whereas the ALiBi variants produce clean, homogeneous results. }
    \label{fig:k_Mmeans}
\end{figure}

The features of DINOv2 have many desirable properties: wide applicability, robustness across `changes of pose, style or even objects' and meaningful unsupervised decompositions (via PCA or $k$-means) \cite{DINOv2}.
We explore some of these properties in \cref{fig:cosine}, \cref{fig:k_Mmeans} and \cref{supp:fig:pca:shared_pca}.

\cref{fig:cosine} shows the cosine similarity for a given query point (red patch) for DINOv2 and our ALiBi models. 
For heterogeneous natural images the results are similar, but for out-of-distribution homogeneous images like an SEM image of a battery cathode ALiBi-Dv2 has a far more homogenous response. 

We show the $k$-means \cite{SKLEARN} based decompositions for the DINOv2, DVT, DINOv3, ALiBi-Dv2 and ALiBi-Dv3 features of a set of natural images in \cref{fig:k_Mmeans}.
Features were scaled (zero-mean, unit standard deviation) channel-wise prior to clustering with 15 initializations, and all 384 output features were used.
For natural images the results are similar, but for some images (bubbles, eutectic steel alloy \cite{DOITPOMS}), DINOv2 and DVT can decompose on positional information, either vertically or radially; as mentioned previously, DVT tends to split vertically.
In contrast, our ALiBi models decompose homogenously. 

Note these examples were selected to demonstrate positional decomposition; for in-distribution heterogeneous images, the features and subsequent decompositions for DINOv2 and especially DVT are generally of high-quality.

\subsection{Benchmarking semantic segmentation}

\begin{table}
    \centering
    \begin{tabular}{l c c c}
    \toprule
        Model &  VOC07 & VOC12 & ADE20k \\
    \midrule
        DINOv2 & 0.680 & 0.762 & \textbf{0.240} \\
        NoPE & 0.689 & 0.757 & 0.235 \\
        \textbf{ALiBi-Dv2} & \textbf{0.692} & \textbf{0.769} & 0.239 \\
    \bottomrule
    \end{tabular}
    \vspace{0.4em}
    \caption{Validation mIoU (mean Intersection over Union) for frozen-feature linear probe semantic segmentation of benchmark datasets. We see our ALiBi-Dv2 performs similarly to, and in some cases outperforms, DINOv2.  }
    \label{tab:sem_seg_lin_probe}
\end{table}

To further assess whether our ALiBi-Dv2 model retains useful DINOv2 semantics, we train linear probes atop frozen ViT features across a series of benchmark segmentation datasets. 
Linear probe training was similar to previous works \cite{DINOv2, DVT}: bilinear upsampling of output features followed by a linear layer and batch norm. 
Note the probes were trained for a relatively short period compared to other works, hence lower mIoUs. The full training setup is described in \cref{supp:sec:sem_seg:training}.

\cref{tab:sem_seg_lin_probe} contains the validation mIoU for linear probes trained on the VOC07 \cite{VOC07}, VOC12 \cite{VOC12} and ADE20K \cite{ADE20K} datasets for DINOv2, NoPE and ALiBi-Dv2.
We see that ALiBi-Dv2 performs \rebuttal{similarly to  DINOv2 and NoPE}, indicating that the model retains both general semantics and the ability to model long-range interactions. Some example predictions are shown in \cref{supp:fig:linear_probe_predictions}.

The \rebuttal{small differences in performance} of ALiBi-Dv2 compared to DINOv2 could be attributed to either improved homogeneity (probes may previously have fit to positional information) or mild artefact removal. 
NoPE performs remarkably well; it seems generally able to construct objects from patch-patch similarity and continuity. \cref{supp:sec:nope_vs_alibi} however contains some examples where this is not the case. 
We also measure the performance of probes under input image transformations such as translations with wrapped boundaries, flips and rotations, this can be found in \cref{supp:tab:sem_seg_transforms}.

\rebuttal{Since our ALiBi models have (minus the learned embedding) the same parameter count as their original counterparts, and are trained to target their embeddings, we do not expect them to possess `better' or more complex features. 
This means their performance in downstream tasks should be upper-bounded by the ViT they are finetuned from - except in cases where the reduction of positional biases is of high importance. We explore one such example in the next section. }

\subsection{Applied to trainable segmentation}
\label{sec:trainable_seg}
\begin{figure*}
\centering
    \includegraphics{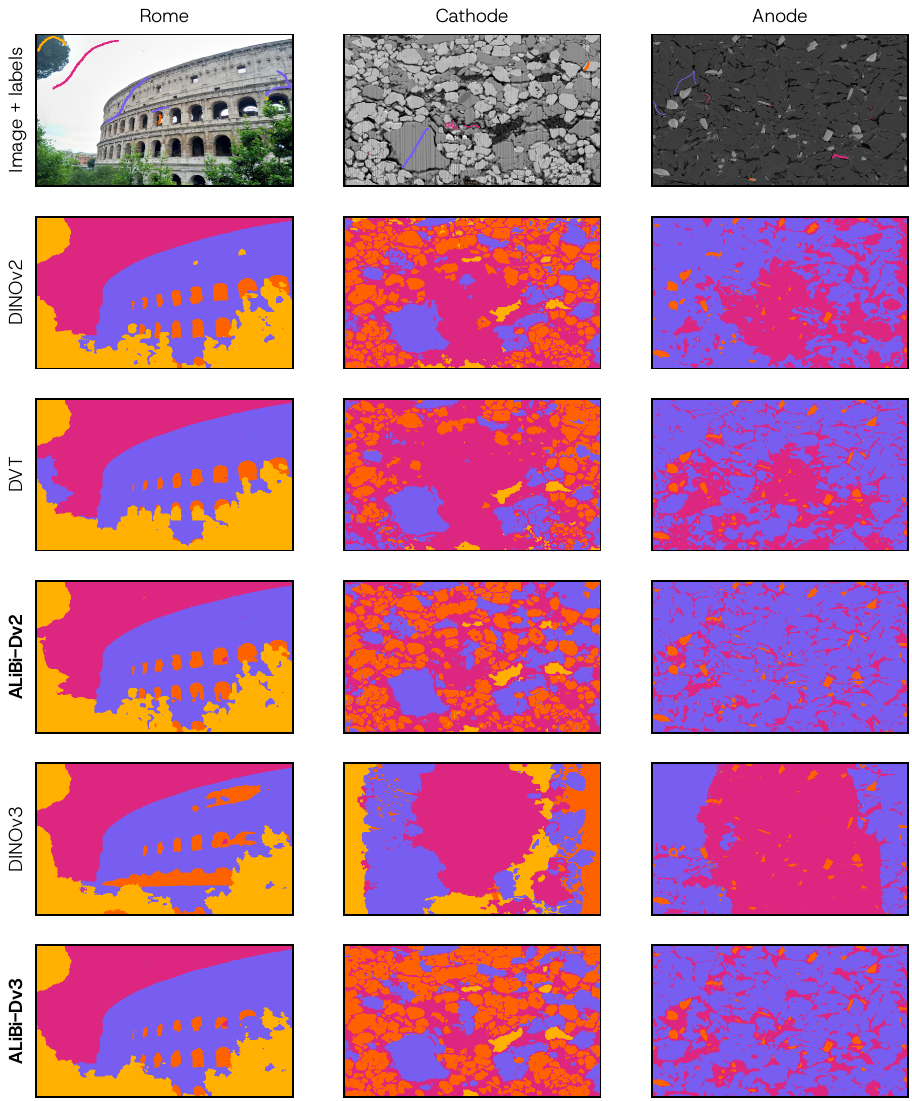}
    \caption{Trainable segmentation of images: a set of features that describe each pixel (classical + bilinearly upsampled ViT features) are extracted, and a lightweight classifier (XGBoost) is trained to map from feature vectors to user labels. Our ALiBi-Dv2 has more homogenous features and so produces less positionally biased (and therefore higher quality) segmentations of complex microstructures. Segmentations using DINOv2 and DVT's features are biased either radially, horizontally or vertically. }
    \label{fig:trainable_seg}
\end{figure*}

One area where more homogeneous ViT features are important is trainable segmentation of materials science micrographs.
The trainable segmentation paradigm is useful for researchers because it allows flexible class definitions, works with very sparse label data and trains quickly \cite{WEKA, ILASTIK}.
These segmentations are necessary to perform quantitive analysis of materials, from basic properties like particle size and distribution \cite{NMC_SEG_ANALYSIS}, to correlating with device performance (yield strength, battery charging speed, degradation) and finally to end-to-end optimisation of the manufacturing workflow \cite{STEVE_MATTER}.

Following Refs. \cite{HR_DV2} and \cite{VULTURE}, we perform trainable segmentation for a set of images using DINOv2, DVT, DINOv3, and ALiBi model features. 
The corresponding ViT features are scaled, reduced to 9 dimensions via PCA and then are bilinearly upsampled to match image dimensions.
Next we concatenate these with the classical features for that image (\textit{i.e.} a collection of multiscale blurs, edge detectors - see \cref{supp:sec:trainable_seg:setup}).
Finally we train an XGBoost classifier \cite{XGBOOST} to map from the features of labelled pixels to their class, followed by conditional random field (CRF) post-processing \cite{CRF}. 
The labels, classical features and classifier parameters were the same for each image - the only difference was the choice of ViT features. The predictions are shown in \cref{fig:trainable_seg}.

There is little difference for the natural image, however the differences for the microstructural images (SEM images of a Li-ion battery cathode \cite{BIPHASE_CATHODE} and silicon-graphite anode \cite{BIL}) are stark.
Segmentations using DINOv2 and DVT features are clearly positionally biased, failing to predict classes either in the centre or bottom of the images.
DINOv2's left-to-right bias and DVT's up-to-down bias (quantified in \cref{sec:results:linear_probe}) are apparent. 
The ALiBi models' predictions are far more homogeneous.

The key difficulty in segmenting these battery materials is capturing the `pore-back' effect \cite{KINTSUGI}.
Researchers typically create flat cross-sections by ablating or milling an area with an ion beam or laser before imaging with their electron microscope.
However, batteries are porous materials (\textit{i.e.} have gaps in their structure - important for transport) and when researchers image cross-sections, closely-located material from further `down' the sample can appear in the same level cross-section image.

The goal is to correctly segment this out-of-plane or `pore-back' material as pore - otherwise statistics become skewed and later simulations non-physical. 
ViT features have previously proved useful at capturing the in-plane vs out-of-plane distinction where classical features previously struggled.
However, if positional information is fit to instead, or is intermixed in the ViT's representation of in- and out- of plane material, then poor segmentation results are obtained.

We note these labels are, in a sense, adversarial in that they are very sparse and spread across the image. 
More labels, placed closer together can improve performance for the other ViT features. 
The `pore-back' effect is just one example of where ViTs can improve materials image analysis - we include further examples in \cref{supp:fig:more_trainable_seg}, as well as a quantitive benchmark in \cref{supp:fig:trainable_seg_bench}. All benefit from the improved homogeneity of the ALiBi models.

\section{\rebuttal{Limitations}}
\label{sec:limiations}
\rebuttal{Despite the success of our finetuning approach in reducing the positional biases of these ViTs, this approach has several limitations to consider.
The first is that there are cases where positional biases may be useful to the problem, such as layered materials, fixed experimental setups, or images that contain spatial gradients.
However, it is possible that these could be added to the features post-hoc (for example, adding a positional grid like ilastik does \cite{ILASTIK}) to account for this.
One of the appeals of ViTs with reduced positional biases is that they enable this decoupling.
}

\rebuttal{There is also an additional memory cost in the forward pass which comes from storing the full dense $N_t \times N_t$ adjacency matrix.
For a $(1500,1500)$ image, the forward pass of a `\texttt{scaled\_dot\_product\_attention}` (SDPA) DINOv2 model uses 240 MB of VRAM, whereas the same SDPA ALiBi model uses 765 MB - this comes entirely from the distance matrix.
This could be addressed in the future via custom kernels.}

\rebuttal{Finally, our specific ALiBi formulation has worse generalisation to images with larger objects.
In \cref{supp:fig:size_extrapolation}, we resize an image from (518, 518) to (1246, 1246) and extract features at each image size, comparing across DINOv2, DINOv3, our ALiBi DINOv2 and our NoPE checkpoint.
DINOv2 and DINOv3 retain semantic coherence - patches belonging to the same object (\textit{i.e.} the head of the dog) retain similar values.
However, for our ALiBi-Dv2 model (and to a greater extent, the NoPE model), the features become more focused on local colour and texture.
This difficulty in modelling large objects may not be a problem in materials microscopy, where the average feature size is small, though there are cases (such as high-resolution imaging) where this could be problematic.
It is possible to mitigate this by fine-tuning with no normalization in the distance matrix, though this increases the positional bias of the resulting model.
}

\section{Conclusion}
\label{sec:conclusion} 

In summary, we have examined positional biases in feature foundation models, finding them present across a range of training objectives and positional encodings. 
These biases tend to be linearly decodable, often manifesting as simple ramps in the output features whose channel-order is consistent across different images.

The difficulty in removing these biases motivated us to train a DINOv2 checkpoint with ALiBi positional encodings. We found that biased ViT embeddings could be used as a training target to produce a homogeneous feature model. Once trained, we showed these models produce rich features that retain the semantics and useful properties of the original ViT (DINOv2 or DINOv3) without the problematic positional biases. This meant they could be applied successfully to trainable segmentation of complex microscopy images that previous models struggled with.

Whilst we found it possible to finetune an existing ViT checkpoint with ALiBi encodings, we did not explore whether it is possible to train a model like DINOv2 from scratch with ALiBi positional encoding. We also found evidence that this positional bias was a more general property of self-supervision (rather than being isolated to the DINO objective or learned positional encodings), but we have yet to explain why. Both questions represent interesting future work.

\section*{Code Availability}
The code needed to reproduce the results of the paper is available at \url{https://github.com/tldr-group/dino-saw} with an MIT license agreement.

\section*{Acknowledgements}
This work was supported by funding from the the EPRSC and SFI Centre for Doctoral Training in Advanced Characterisation of Materials (EP/S023259/1 received by RD), the Royal Society (IF\textbackslash R2\textbackslash 222059 received by AV as a Royal Society Industry Fellow) and the EPSRC Open Plus Fellowship "AIMS-DEEP" (Award UKRI4070 received by SJC).

AB, AW and MP thank the German Ministry for Research, Technology and Space (BMFTR) for funding the SiMba project (03XP0584).

The authors would like to thank other members of the TLDR group for their testing and feedback.

\section*{Authorship}
MP, SJC, AV and RD conceptualised the project. MP and RD wrote the code and performed the experiments. AV, AW, AB and SJC contributed to the development of the concepts presented in the work, and provided funding \& equipment access. AV, AW, AB, SJC and RD provided supervision during the project. All authors contributed to revisions and edits of the manuscript.

\section*{Competing interests}
The authors declare no competing interests.

\section*{References}
\addcontentsline{toc}{section}{References}

\def\addvspace#1{}

	\renewcommand{\refname}{ \vspace{-\baselineskip}\vspace{-1.1mm} }
	\bibliographystyle{ieeetr}
    \bibliography{main}


\onecolumn
\section*{Supplementary}
\setcounter{section}{0}
\setcounter{figure}{0}
\setcounter{table}{0}
\renewcommand*{\theHsection}{S.\the\value{section}}

\makeatletter
\renewcommand \thesection{S\@arabic\c@section}
\renewcommand\thetable{S\@arabic\c@table}
\renewcommand \thefigure{S\@arabic\c@figure}
\makeatother


\section{Linear probe details}
\subsection{Dataset \rebuttal{details}}
\label{supp:sec:linear_probe:dataset}

\begin{figure}
\centering
    \includegraphics{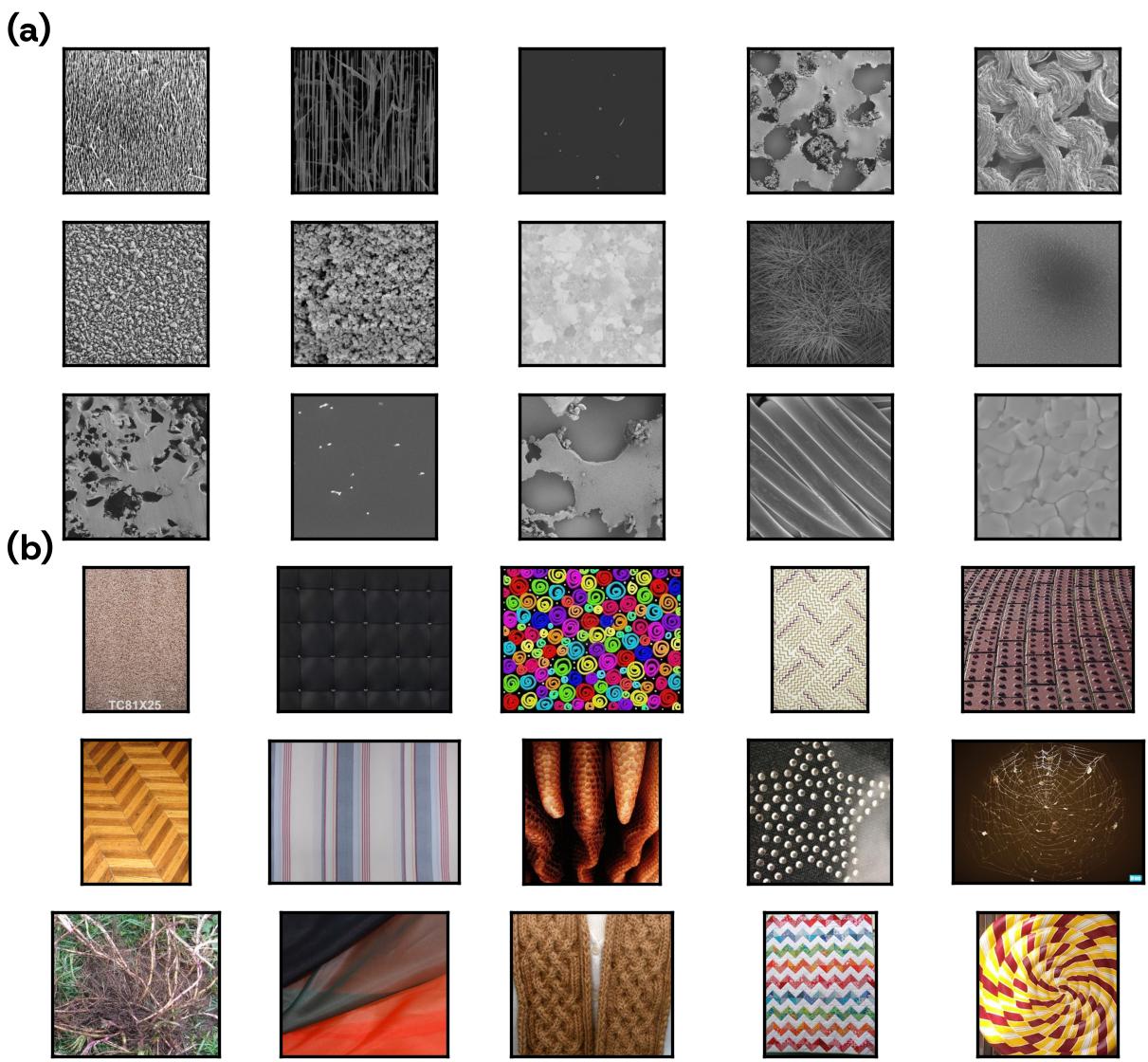}
    \caption{\textbf{(a)} \rebuttal{Random samples from the micrograph dataset used for the linear probe.} \textbf{(b)} Random samples from the DTD used for the linear probe \cite{DTD}.}
    \label{supp:fig:linear_probe_ds}
\end{figure}

We show examples for the linear probe datasets in \cref{supp:fig:linear_probe_ds}. Note that some images from the texture dataset \cite{DTD} are not perfectly homogenous, having either spatially biased content, or colour / contrast / depth gradients. These could be fit to in principle, and may be why our ALiBi-Dv2 has a higher $R^2$ score on DTD compared to random noise or micrographs.

\rebuttal{For the micrograph dataset, we took the 22,000 SEM images from \cite{BIG_SEM_DATASET}, chose 6 of the most homogeneous categories (these being "Powder", "Fibres", "Particles", "Nanowires", "Porous Sponge", "Films Coated Surface", and randomly chose 100 images, sampling equally from each category and resulting in $\approx$ 16 images per category.
We performed longest-side resizing to 1024 px, and then centre-cropped to the length of the shortest side to maintain a square aspect ratio.
This resulted in a set of images between 600 - 800 px.}

\subsection{Sampling strategies}

To check the influence of different sampling methods, we compare different strategies: grid-based interpolation, grid-based interpolation with holdout regions for extrapolation, and random sampling.

No significant differences between the different sampling strategies are observed.
Only minor deviations appear for the grid\_holdout setting compared to the others. This may be explained by the presence of the holdout region, which provides less information about the ramp compared to the other approaches.

\begin{figure}
\centering
    \includegraphics{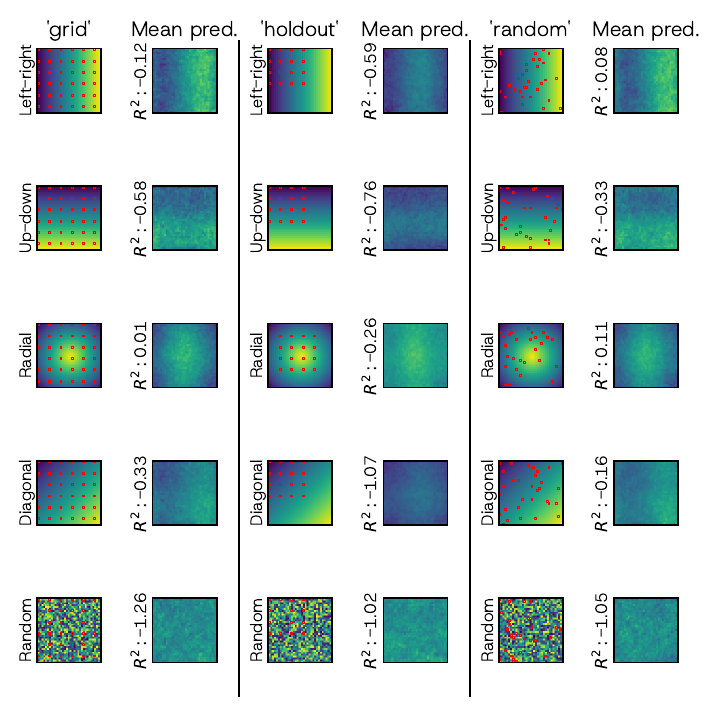}
    \caption{Comparison of sampling strategies. For each method (grid, holdout, random), the target ramps (left) and the mean model predictions across all channels (right) are shown. Red markers denote sampled training locations.}
    \label{fig:lin_data_sampling}
\end{figure}

\subsection{Random ramp results}

\begin{figure}
\centering
    \includegraphics{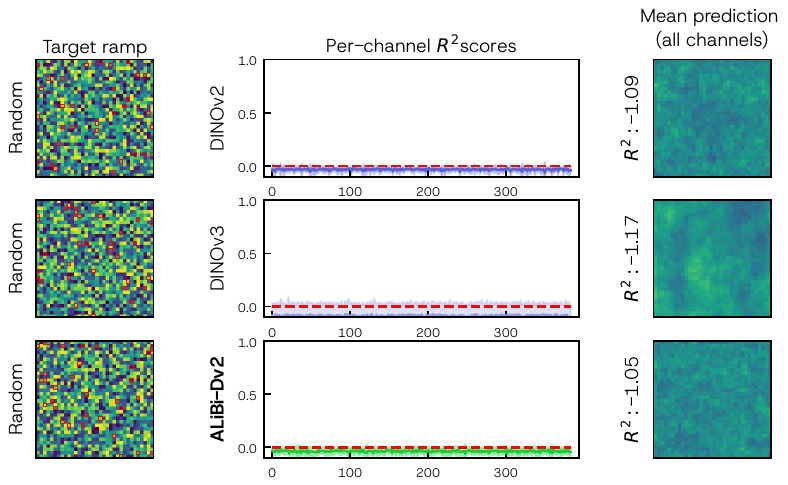}
    \caption{Linear probe analysis of DINOv2, DINOv3 and ALiBi-Dv2 with a target ramp of random noise, using the micrograph dataset.}
    \label{supp:fig:random_sampling}
\end{figure}

To confirm our linear probing is not fitting to spurious correlations in the features, we attempt to fit to a random noise ramp, plotting the results in \cref{supp:fig:random_sampling}. We see low $R^2$ scores for all models.

\subsection{Results on additional models}
\label{supp:sec:linear_probe:more_models}
\begin{figure}
\centering
    \includegraphics{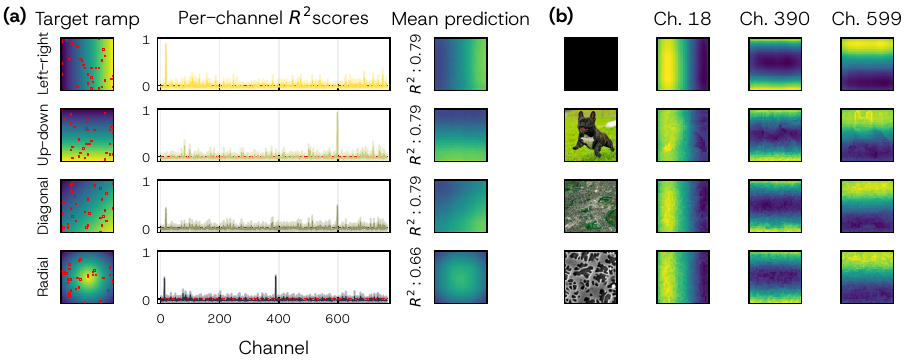}
    \caption{Linear probe analysis of ViT-B (MAE) features over the micrograph dataset. Again we see clear ramp channels in the output features.}
    \label{supp:fig:vit_b_probe}
\end{figure}

\begin{figure}
\centering
    \includegraphics{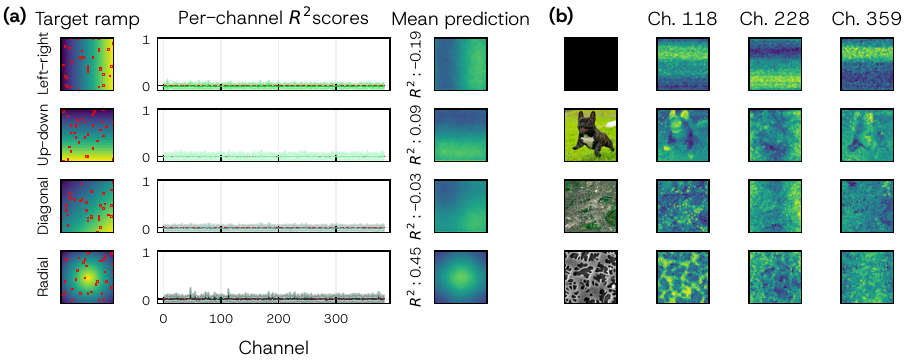}
    \caption{Linear probe analysis of our ALiBi-Dv2 features over the micrograph dataset.}
    \label{supp:fig:alibi_probe}
\end{figure}

We present linear probe analysis of ViT features for additional models in \cref{supp:fig:vit_b_probe} and \cref{supp:fig:alibi_probe} (ViT-B MAE \& our ALiBi-Dv2 respectively). 
Again we note highly-positional (ramp-like) outlier channels, indicating this phenomenon is not limited to the DINO training objective.
Our ALiBi-Dv2 model does not show these outlier channels, and its mean prediction scores (when training on all channels) are consistently low.

\subsection{Layer channel-behaviour of models}
\label{supp:sec:linear_probe:layer_channel_behaviour}

\begin{figure}
\centering
    \includegraphics{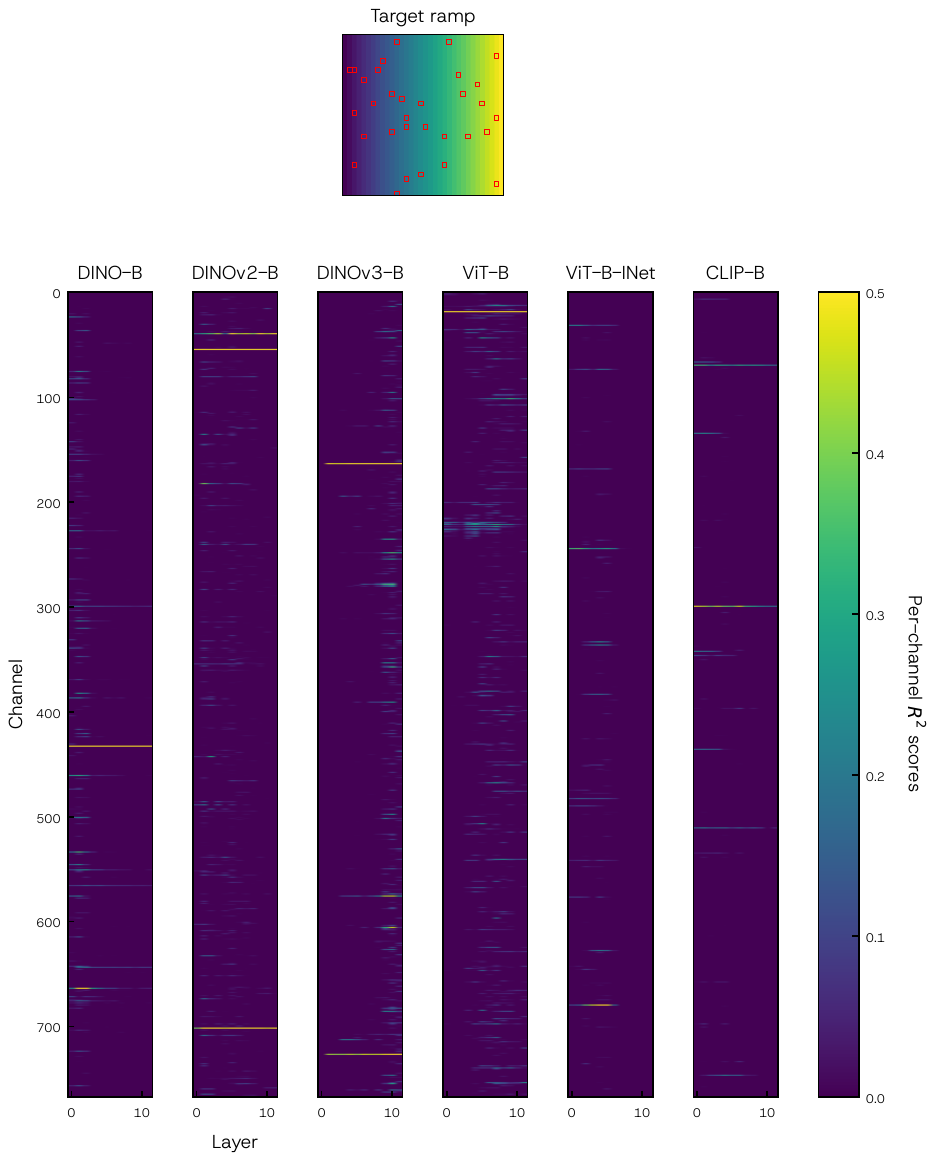}
    \caption{Per-channel per-layer $R^2$ scores for an $(x,y)$ ramp for a series of models. We see self-supervised models (DINO-B, DINOv2-B, DINOv3-B, ViT-B) have stronger and more widespread positional channels.}
    \label{supp:fig:fingerprints_diff_models}
\end{figure}

We performed channel-wise linear probing (targeting an $(x,y)$ ramp) as a function of layer for a series of literature ViT backbones, presenting the result in \cref{supp:fig:fingerprints_diff_models}.

We note two things: first, SSL backbones have stronger and more widespread positional information compared to supervised models (ViT-B-INet, CLIP-B) and second that DINOv3 becomes more positional as layers increase, whereas DINO and DINOv2 become less positional.
DINO and DINOv2 use learned positional encodings, added at the start of the model, whereas DINOv3 uses RoPE, which is applied at each attention layer.

\subsection{Cross-image probing}
\begin{table*}
    \centering
    \begin{tabular}{l c c c c c}
    \toprule
        Model & \textit{d} & PE & $R^2_{\text{A} \to \text{A (micro)}}$ & $R^2_{\text{A} \to \text{B (micro)}}$ & $R^2_{\text{A} \to \text{B (texture)}}$ \\
    \midrule
        DINOv2 & S & Learned & \cellcolor{red!25} 0.80 & \cellcolor{red!25} 0.59 & \cellcolor{red!25} 0.49 \\
        \textbf{ALiBi-Dv2} & \textbf{S} & \textbf{ALiBI} & \cellcolor{green!25} \textbf{0.00} & \cellcolor{green!25} \textbf{-0.71} & \cellcolor{green!25} \textbf{-0.56} \\
    \bottomrule
    \end{tabular}
    \caption{\rebuttal{$(x, y)$ linear probe $R^2$ scores trained on micrograph images (A) and evaluated on: held-out patches of the same image ($\text{A} \to \text{A}$, `micro'), patches of another micrograph image ($\text{A} \to \text{B}$, `micro'), and patches of a texture image ($\text{A} \to \text{B}$, `texture'). We see some biases transfer across images - such as the pathological channels identified in \cref{sec:results:linear_probe}.}}
    \label{tab:cross_image_linear_probe}
\end{table*}

\rebuttal{As a further test of these positional biases, we performed the following experiment. For a linear probe trained on a random micrograph from the dataset, A, we applied it to either the same micrograph (\textit{i.e.} the same as \cref{sec:results:linear_probe}), A (micro), a different random micrograph, B (micro), or random image from the texture dataset, B (texture).
We did this for both DINOv2 and our ALiBi-Dv2, averaging results over 100 random samples, and targeting an $(x,y)$ ramp. Other settings (train/test fraction, random sampling) were kept from \cref{sec:method:lin_probe}.
The scores are reported in \cref{tab:cross_image_linear_probe}, note we are reporting scores on holdout test patches.
}

\rebuttal{
A significant positional bias remains decodable across images - we attribute this partly to the persistent channels we identified in \cref{sec:results:linear_probe}.
These biases also remain decodable across domains (micrographs $\rightarrow$ textures), though again with a slight decrease.
Our ALiBI-Dv2 checkpoint does not suffer from these biases, and sees a large decrease in $R^2$ when applying to different images.
}

\section{Model details}

\subsection{Hyperparameters}
\label{sec:supp:model:hyperparams}

In \cref{tab:hyperparams} we list the hyperparameters used to train the ALiBi-Dv2 model, both for high- and low-resolution setup.
Both configurations use the same optimizer and cosine similarity loss, while resolution, learning rate, batch size, and number of epochs differ.

\begin{table}
    \centering
    \begin{tabular}{l c c}
    \toprule
        ALiBi-Dv2 & low resolution & high resolution \\
    \midrule
        image size & (224, 224) & (518, 518) \\
        learning rate & 1e-4 & 1e-5 \\
        batch size & 256 & 32 \\
        num epochs & 15 & 5 \\
        optimizer & AdamW ($\lambda=0.01$) & AdamW ($\lambda=0.01$) \\
        loss & cosine similarity & cosine similarity \\
    \bottomrule
    \end{tabular}
    \vspace{0.4em}
    \caption{Training hyperparameters for ALiBi-Dv2 in the low- and high-resolution training setups.}
    \label{tab:hyperparams}
\end{table}

\subsection{Trainable vs constant ALiBi heads}
\label{sec:supp:model:trainable_m}
\begin{figure}
\centering
    \includegraphics{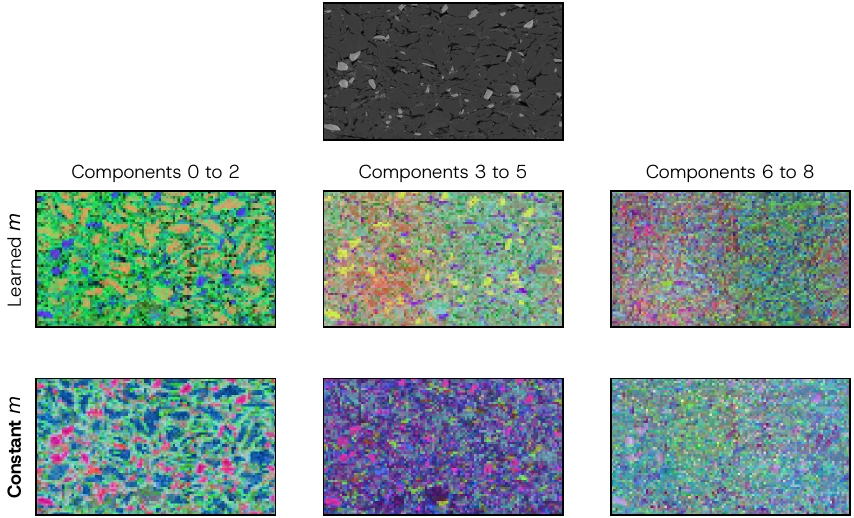}
    \caption{Comparing the impact of learned per-head ALiBi scalars \textit{m} (top row) and constant \textit{m} $1$ (bottom row). We see learning the head scalars can lead to undesirable long-range positional biases.}
    \label{supp:fig:train_v_const}
\end{figure}

In \cref{supp:fig:train_v_const}, we compare training the ALiBi distance matrix scaling parameter \textit{m} with keeping it fixed at a value of $1$. Fixing the parameter results in qualitatively more homogeneous PCA compared to the model in which \textit{m} is learned during fine-tuning.
We hypothesize that this arises because heads learn distinct length scales, thereby gaining the capacity to reconstruct positional biases across multiple layers.

\subsection{Multiscale training}
\label{sec:supp:model:multiscale_training}

\begin{figure}
\centering
    \includegraphics[width=0.75\linewidth]{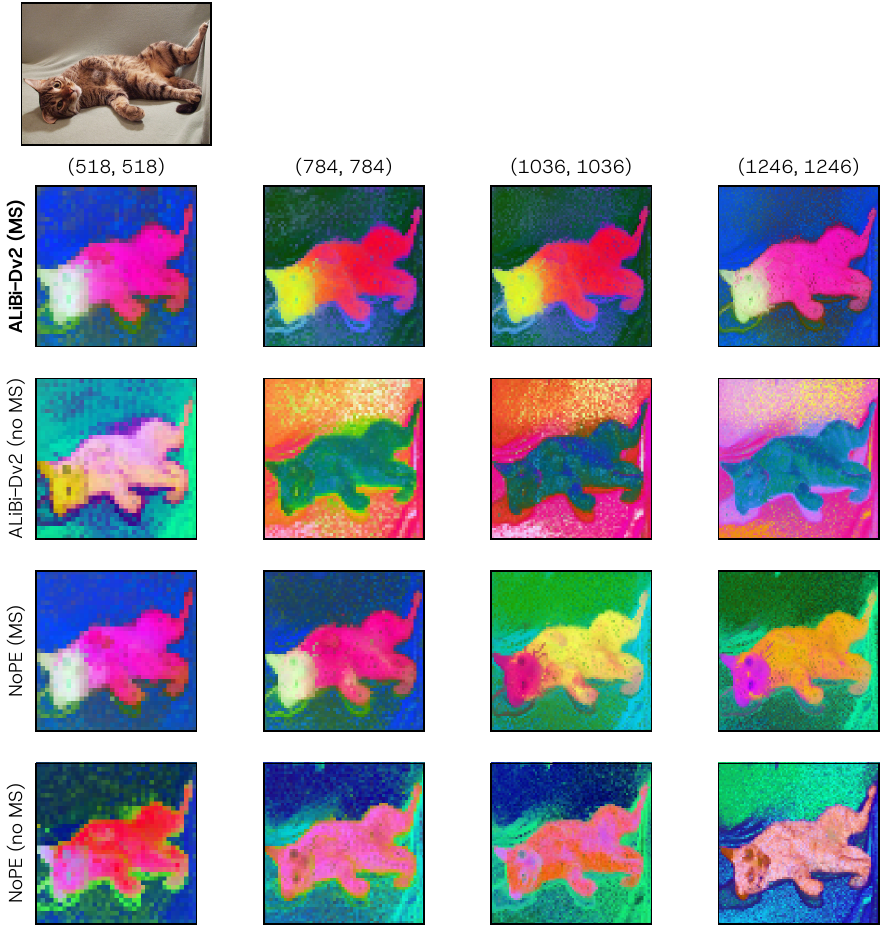}
    \caption{
        PCA visualizations of patch embeddings demonstrate that models trained with multiscale augmentation (ALiBi-Dv2, NoPE) successfully distinguish semantic regions (\textit{e.g.} separating the head from the body) across varying input resolutions. Models lacking multiscale training (-no-ms) fail to separate these features, highlighting the necessity of multiscale training for adapting to varying feature scales.
    }
    \label{supp:fig:multi_scale_training}
\end{figure}

We found training for a short period on larger images during finetuning results in better features for larger images (similar to \cite{DINOv2, DINOv3}).
When images are simply resized to higher resolutions, the effective scale of visual features increases while the patch embedding continues to operate on fixed (14,14) pixel patches, leading to a mismatch in feature scales. 
This difference becomes particularly apparent in \cref{supp:fig:multi_scale_training}: models trained with a multiscale step clearly distinguish the cat’s head from its body, whereas models trained without multiscale augmentation fail to separate these regions.
We see multiscale training also improves NoPE - this is further indication that multiscale training is largely about accommodating the patch-image feature size ratio, rather than for to adjust to new (interpolated) positional encoding values.

\section{Sinusoid vs NoPE vs ALiBi}
\label{supp:sec:nope_vs_alibi}

\begin{figure}
\centering
    \includegraphics{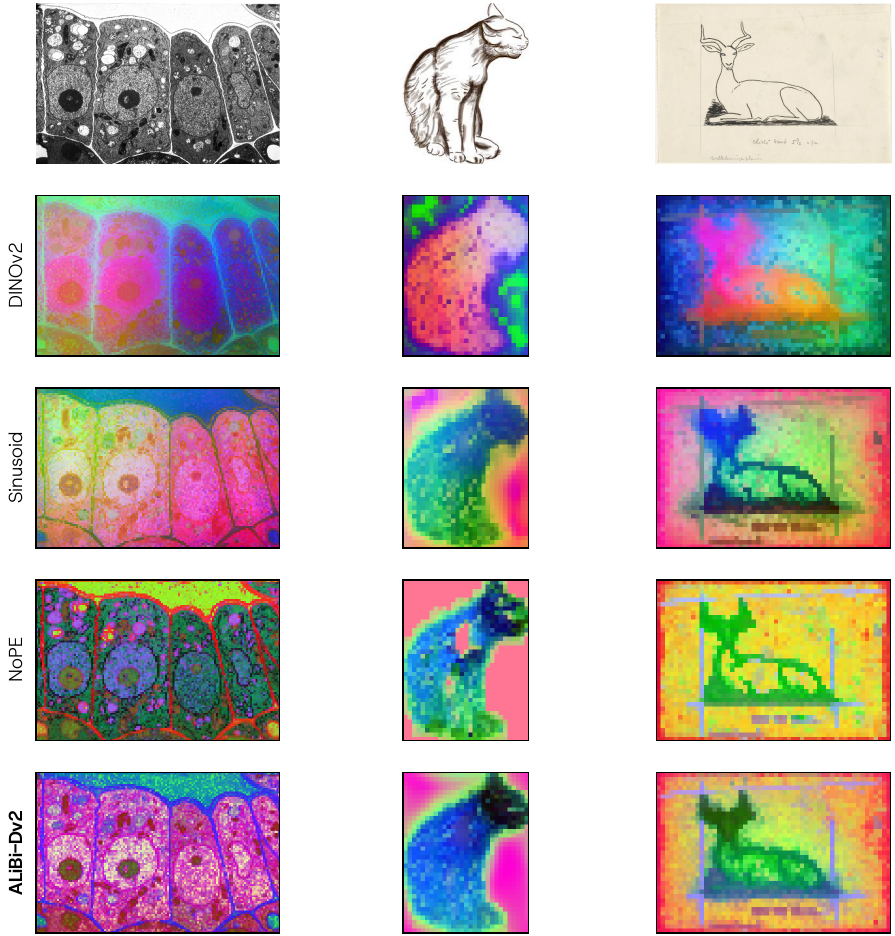}
    \caption{Feature PCAs with different finetuned positional encodings. When using sinusoidal PE, positional biases re-emerge during training (even with channel-blanking); NoPE has no biases (by definition), but fails to model useful spatial relationships for similar patches (see the shoulder of the cat). ALiBi-Dv2 avoids these problems.  }
    \label{supp:fig:nope_v_alibi}
\end{figure}

\cref{supp:fig:nope_v_alibi} shows a comparison of feature PCAs for different feature models, specifically DINOv2, a DINOv2 variant finetuned with a fixed sinusoidal PE, a DINOv2 variant finetuned with NoPE (\textit{i.e.} bag of patches), and our ALiBi-Dv2.
The sinusoidal DINOv2 retains the positional biases, despite being finetuned with channel blanking - this is because its fixed absolute PE can easily model the biases present in the DINOv2 features.
\rebuttal{NoPE, on the other hand, is incapable of modelling these problematic biases, but then consequently struggles to capture relationships in the cases where there are similar patches across different phases or objects.
For example, the organelles of the cell and the shoulder of the cat are both pure white, the same as the background, and so are indistinguishable to NoPE.
Our ALiBi-Dv2 model, by virtue of being able to express relative distances, suffers less from this problem.
}

\rebuttal{
Although we are often segmenting homogeneous materials - and therefore do not care if patch A of active material has a different representation to patch B -  there are many cases where identical patches can belong to different phases, and are only distinguishable through wider spatial context.
For example, an empty patch in a wide pore space of a Li-ion cathode vs an empty patch inside a large crack in an active material particle, or the single-crystal NMC vs. the agglomerate polycrystal.
We would like our deep features to be able to distinguish between these cases (and still be relatively positionally unbiased) - this is where our ALiBi models are useful.
}

\section{Bias of the zero tensor}
\label{supp:sec:bias_zero_tensor}

\begin{figure}
\centering
    \includegraphics[width=\linewidth]{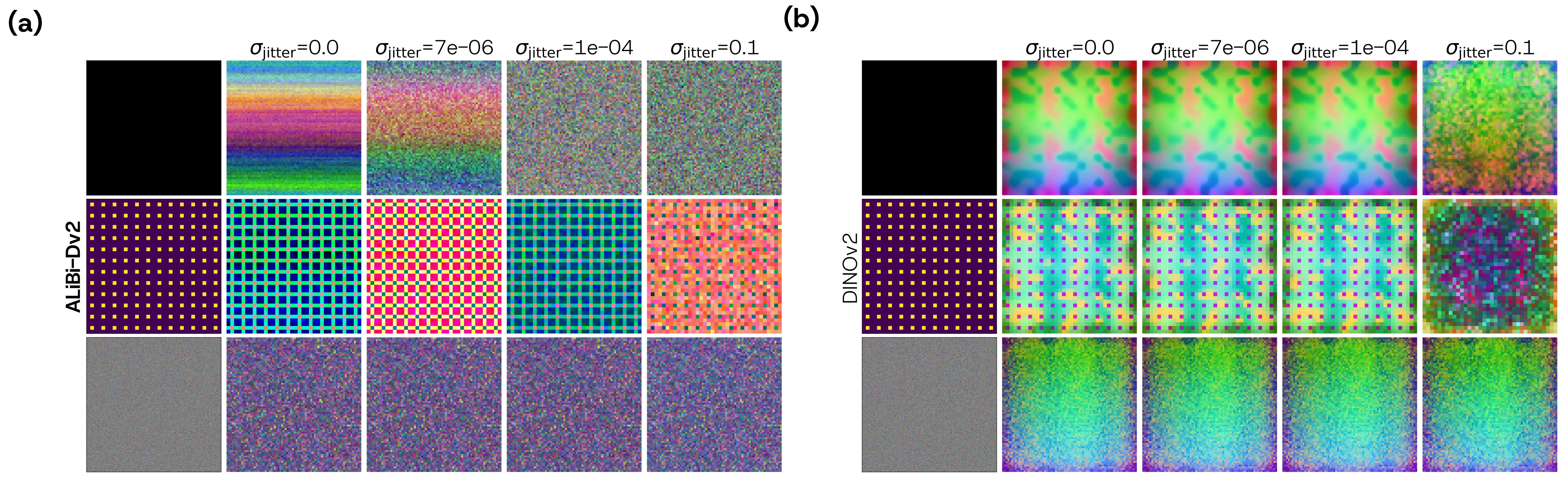}
    \caption{
        PCA of very low information images for \textbf{(a)} ALiBi-Dv2 and \textbf{(b)} DINOv2.
        with introduced random noise into the respective positional encodings.
    }
    \label{supp:fig:zero_tensor_bias}
\end{figure}

Of note in \cref{supp:fig:alibi_probe} is that there appears to be structure in the ALiBI-Dv2 features for an empty image.
To further test this, we plot feature PCAs of low information images - namely the zeros tensor, a uniform grid of squares (of length patch size) and random noise - showing the results in \cref{supp:fig:zero_tensor_bias}. 
For the zeros tensor we see clear bar artefacts, and for the dot grid we see background inhomogeneities. 
For the random noise we see no structure. 

We hypothesised this was due to compounding numerical imprecision (plus layer scales) in the model and our relative distance matrix.
To test this, we add test-time jitter \textit{i.e.} noise drawn $\mathcal{N}(0, \sigma_{\text{jitter}})$ to the distance matrix at each layer. 
As $\sigma_{\text{jitter}}$ increases, the structure disappears.

We compare this to the feature PCAs for DINOv2, adding the noise to the input learned positional encoding (rather than ALiBi's per-layer distance matrix). The structure persists more strongly than for ALiBi-Dv2. For DINOv2 there is structure for the random noise input.

\section{Input size extrapolation}

\begin{figure}
\centering
    \includegraphics{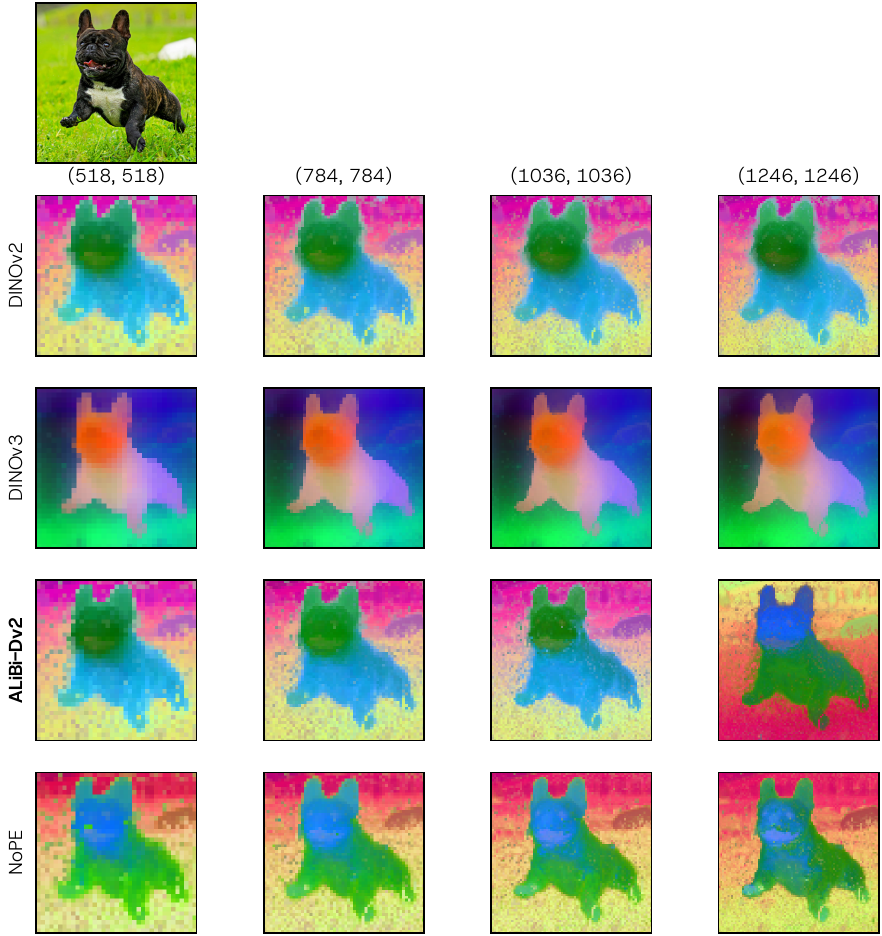}
    \caption{
        PCA features for a series of models as a function of input size resolution. DINOv2 and DINOv3 retain semantics as the image is resized at all resolutions. ALiBi-Dv2 handles resizing well, but loses semantic coherence at $(1246, 1246)$ \textit{i.e.} around 2.5 times the largest image seen during training.
    }
    \label{supp:fig:size_extrapolation}
\end{figure}

Scaling to larger image resolutions highlights the varying generalization capabilities of the models.
While DINOv2 and DINOv3 generalize well, our proposed ALiBi-Dv2 model's features degrade at high resolutions. 
However, although it falls short of the highly robust scaling inherent to the DINO models, ALiBi-Dv2 still yields slightly more coherent and structurally meaningful representations compared to NoPE at these higher resolutions as seen in \cref{supp:fig:size_extrapolation}.

\section{PCA visualisations}
\label{supp:sec:pca}

\begin{figure}
\centering
    \includegraphics{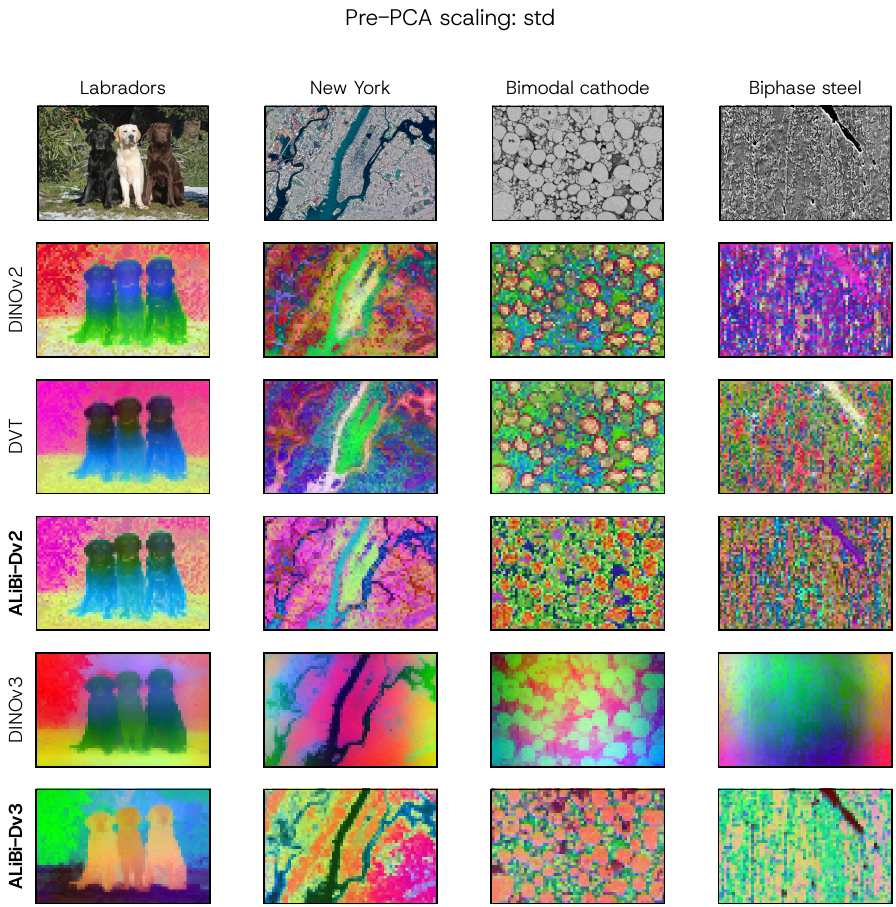}
    \caption{Feature PCAs with feature scaling before PCA. Although the positional biases are less strongly expressed in the visualisation, they are still present, and can still be fit to or clustered around during downstream tasks. }
    \label{supp:fig:pca:more_pca_std}
\end{figure}

\subsection{Shared PCAs}
\label{supp:sec:pca:shared_pca}
\begin{figure}
\centering
    \includegraphics{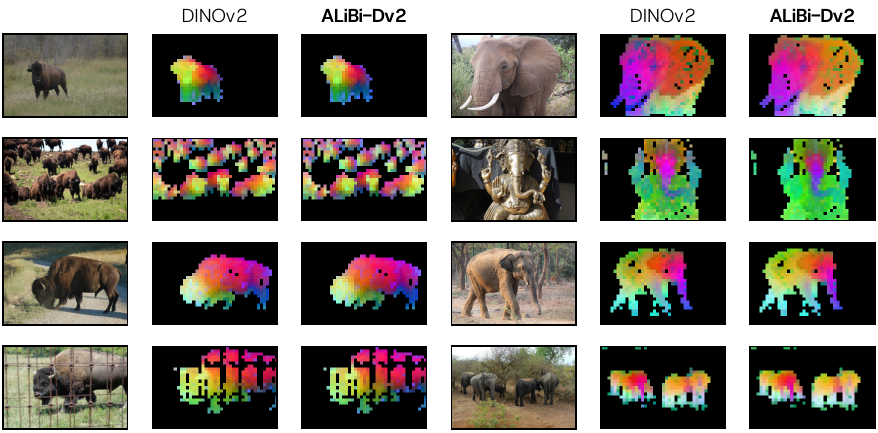}
    \caption{Shared feature PCAs for DINOv2 and ALiBi-Dv2. We see ALiBi-Dv2 retains DINOv2's feature similarities across changes in style, pose, \textit{etc.}}
    \label{supp:fig:pca:shared_pca}
\end{figure}

\cref{supp:fig:pca:shared_pca} shows a shared PCA of foreground features
for DINOv2 and our ALiBi-DV2. There is good agreement, regardless of position, size, orientation or number. This suggests the replacement of the learned PE has not damaged semantics.

\subsection{The perils of PCA visualisation}
\label{supp:sec:pca:pca_perils}

\cref{fig:more_pcas} and \cref{supp:fig:pca:more_pca_std} show the same features for the same images from the same ViT backbones but with no feature scaling and with standard scaling (zero mean, unit standard deviation) before fitting the PCA.
Scaling the features beforehand reduces how strongly the positional biases appear. 
This effect tends to be more apparent for out-of-distribution images (\textit{i.e.} micrographs) - this is understandable, ViT features are naturally less expressive (lower variance and magnitude) for out-of-distribution data, and the postional biases then appear more strongly. 
It is worth noting that the `correct' choice of feature scaling could hide these biases, which would still be present even if not shown.

\section{Segmentation linear probes}
\subsection{Training}
\label{supp:sec:sem_seg:training}

The training procedure largely follows that of \cite{DINOv2, DVT}. We did not perform a grid search and instead fixed the learning rate at 1e-3. Images were resized to (518,518) to match the patch size used in DINOv2. Since ADE20K contains substantially more classes than VOC12 or VOC07, the batch size had to be reduced. Due to computational constraints, the number of training epochs was lower than those reported in \cite{DINOv2, DVT}; however, we expect the general trends to remain consistent. For ADE20K, the 1k subsamples were randomly drawn from the full set of 20k training images in order to improve the logging of loss and mIoU during training.

\begin{table}
    \centering
    \begin{tabular}{l c c c}
    \toprule
        & VOC07 & VOC12 & ADE20K \\
    \midrule
        image size & (518, 518) & (518, 518) & (518, 518) \\
        learning rate & 1e-3 & 1e-3 & 1e-3 \\
        batch size & 64 & 64 & 32 \\
        num epochs & 75 & 50 & 50 \\
        samples per epoch & 209 & 1,4k & 1k \\
        optimizer & AdamW($\lambda=0.01$) & AdamW($\lambda=0.01$) & AdamW($\lambda=0.01$) \\
        loss & cross entropy & cross entropy & cross entropy \\
    \bottomrule
    \end{tabular}
    \vspace{0.4em}
    \caption{Training hyperparameters of the linear probe setup}
    \label{tab:lin_probe_train_params}
\end{table}

\subsection{Transformation robustness}
\label{supp:sec:sem_seg:transform_robustness}

To quantify how robust the models and trained probes are to transformations, we perform the following experiment: transform an input image (flip-ud, roll, rot90), extract features from the ViT, apply the inverse transformation to the features and pass through the probe.
Results are shown in Table \ref{supp:tab:sem_seg_transforms}. 
ALiBi-Dv2 is more robust to adversarial transformations, especially \texttt{roll} - thanks to wrap conditions in the ALiBi distance matrix it is effectively equivariant. 
DINOv2 is relatively robust to the roll transformation, which implies the linear probes were largely ignoring positional biases during training. 
We generally expect this to be the case - when fitting downstream head networks to a large enough dataset these positional biases should be averaged away. 
The problem arises when applying these models in low- or no-data regimes, such as materials science.

\begin{table}
    \centering
    \begin{tabular}{l c c c c}
    \toprule
        Model & \texttt{none} &  \texttt{flip-ud} & \texttt{roll} & \texttt{rot90} \\
    \midrule
        DINOv2&0.762&0.4759&0.7601&\textbf{0.456} \\
        NoPE&0.757&0.482&0.757&0.436 \\
        \textbf{ALiBi-Dv2}&\textbf{0.769}&\textbf{0.499}&\textbf{0.769}&0.445 \\
    \bottomrule
    \end{tabular}
    \vspace{0.4em}
    \caption{Validation mIoU for frozen-feature linear probe semantic segmentation of VOC12 as a function of image transform. ALiBi-Dv2 is generally more robust to transformations than DINOv2 or NoPE. }
    \label{supp:tab:sem_seg_transforms}
\end{table}

\subsection{Linear probe predictions}
\label{supp:sec:sem_seg:preds}
\begin{figure}
\centering
    \includegraphics[width=\linewidth]{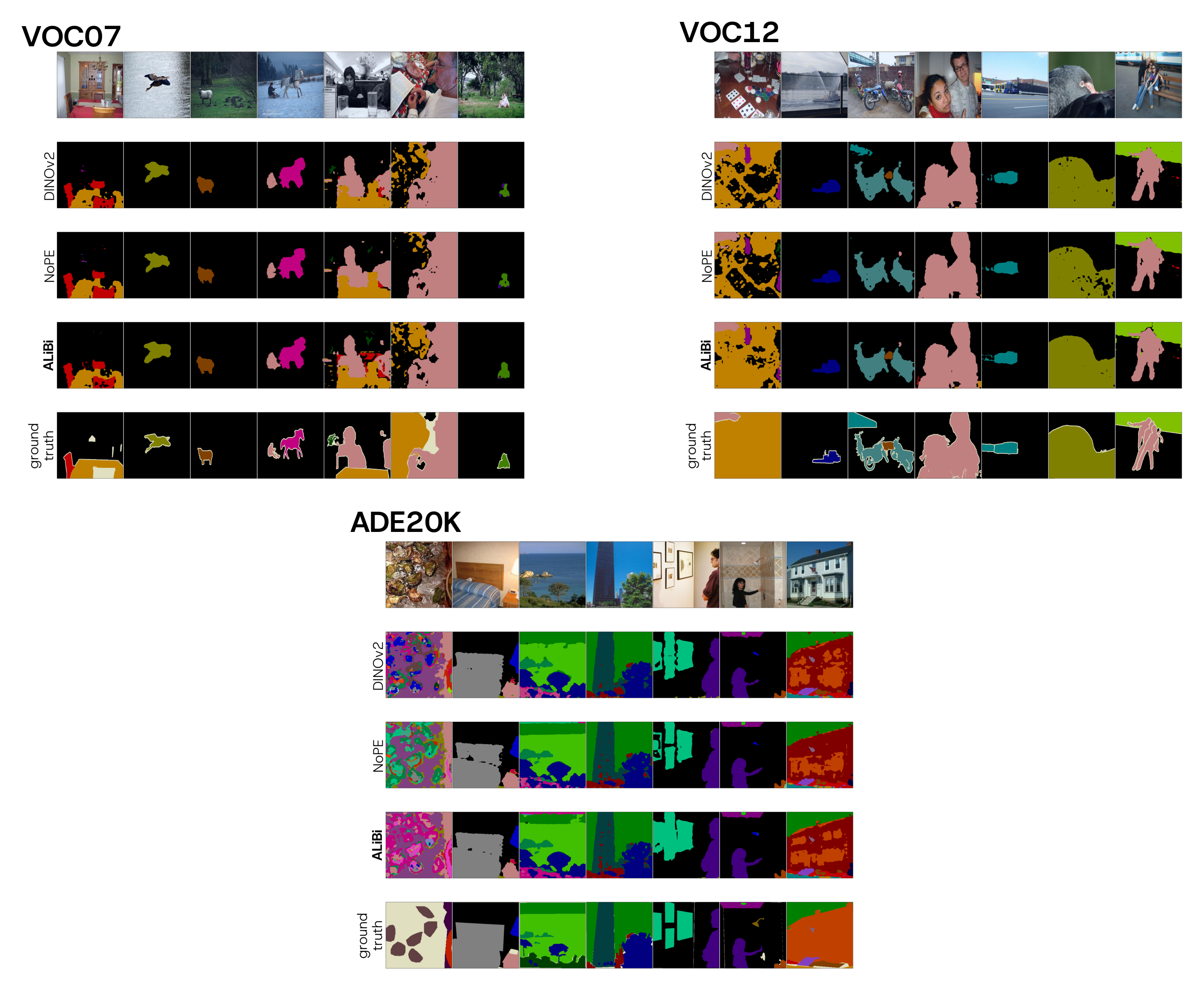}
    \caption{Visual comparison of linear head segmentation predictions using various backbones (DINOv2, NoPE, ALiBi) evaluated on the VOC07, VOC12, and ADE20K benchmarks}
    \label{supp:fig:linear_probe_predictions}
\end{figure}

Figure \cref{supp:fig:linear_probe_predictions} provides additional qualitative examples of linear head predictions across the VOC07, VOC12, and ADE20K benchmarks.
They are qualitatively all very similar due to the results shown in \cref{tab:sem_seg_lin_probe}.
But with minor improvements in VOC07 and VOC12 by ALiBi-Dv2 over DINOv2.

\section{Trainable segmentation}
\label{supp:sec:trainable_seg}
\subsection{Setup}
\label{supp:sec:trainable_seg:setup}
We used the same classical features as Refs. \cite{VULTURE, HR_DV2, WEKA}, namely:
\begin{itemize}
    \item A set of Gaussian blurs of the image with $\sigma \in \{0, 1, 2, 4, 8, 16\}$.
    \item Sobel filter and Hessian filter (plus eigenvalues) on each of these Gaussian blurred images.
    \item Difference of Gaussians across the blurs.
    \item Membrane projections (convolution with rotated line kernels) of the image.
\end{itemize}

We use an XGBoost classifer\cite{XGBOOST}, and CRF post-processing \cite{CRF}.

\subsection{More examples}
\label{supp:sec:trainable_seg:more_examples}

\begin{figure}
\centering
    \includegraphics{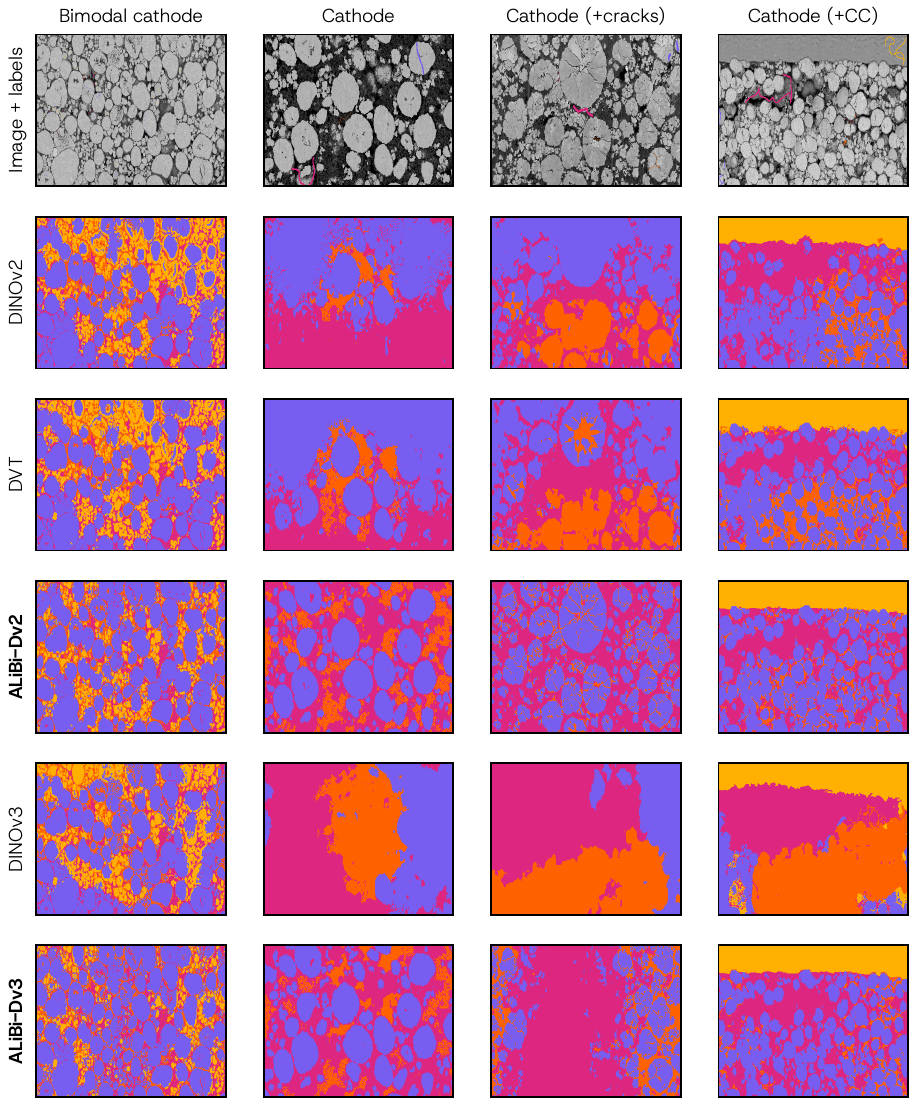}
    \caption{Further examples of trainable segmentation of battery materials using different ViT features. Data from Refs. \cite{BIL}, \cite{KINTSUGI}, \cite{EXPERTSEG} and \cite{BIL}, respectively.}
    \label{supp:fig:more_trainable_seg}
\end{figure}

We present additional examples of trainable segmentation of battery materials in \cref{supp:fig:more_trainable_seg}: a bimodal cathode (containing two different particle sizes for more efficient packing), a calendared (\textit{i.e.} not compressed during manufacturing) cathode, a degraded cathode with cracking, and a cathode with the current collector included in the image. Again we see that positional biases express themselves, except in the case of our ALiBi-Dv2.

Note these are all nominally the same material: a Lithium-Nickel Manganese Cobalt oxide (NMC) cathode - a popular battery chemistry, especially for electric vehicles. Despite this the images vary quite considerably, due to manufacturing parameters or imaging choices.
This wide variation in similar materials is what motivates low or no-training pipelines (such as trainable segmentation),

\subsection{Generalization \& quantitive benchmarks}
\label{supp:sec:trainable_seg:quantitive_bench}
To quantify the improvement our ALiBi-Dv2 adds to materials trainable segmentation, we follow the experimental setup of Ref. \cite{VULTURE}: training classifiers on a sparsely-labelled subset of an image-ground truth micrograph dataset (SEM images of nickel superalloys from \cite{NI_SUPERALLOY}) and measuring mIoU across test images.

We replace their (dense) human labels with algorithmically generated scribbles, in order to measure the performance as a function of added labels. We train on 5 images and apply to a set of 17 unlabelled test images. Similar to \cref{sec:trainable_seg} we use an XGBoost classifier and combine classical with `deep' ViT features. Note we do not use a CRF. We apply labels in a `round', where one round = one scribble label per-class per-training image (\textit{i.e.} 3 classes $\times$ 5 images = 15 labels per round). Labels are always correct, in that they respect the ground truth.

\begin{figure}
\centering
    \includegraphics[width=\linewidth]{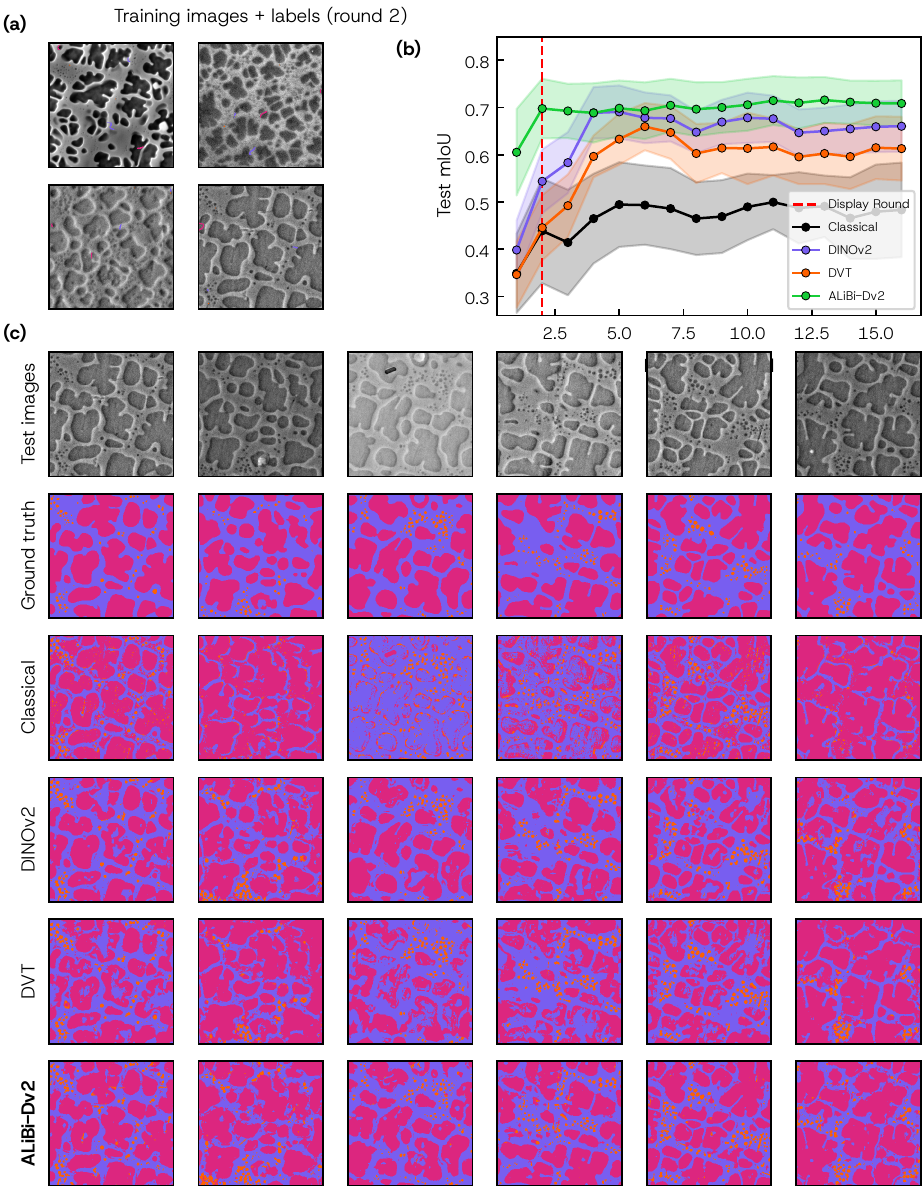}
    \caption{Trainable segmentation quantitive benchmark. \textbf{(a)} For a set of 5 training images (\rebuttal{only four are shown}), N rounds of human-like labels were added. \textbf{(b)} The class-averaged mIoU over 17 test images (with no labels) as a function of number of labelling rounds. \textbf{(c)} Example predictions of XGBoost classifier trained to map features $\rightarrow$ labels after 5 labelling rounds for classical, classical + DINOv2, \rebuttal{classical + DVT}, and classical + ALiBi-Dv2. ALiBi-Dv2 offers improved performance over DINOv2 with less labels due to the lack of positional bias in its features.}
    \label{supp:fig:trainable_seg_bench}
\end{figure}

The results of this experiment are plotted in \cref{supp:fig:trainable_seg_bench}. Similar to Ref. \cite{VULTURE}, we find the addition of ViT features improves the mIoU drastically (around 0.3 points), as the classical features struggle with the variations in exposure and noise. ALiBi-Dv2 improves the mIoU compared to DINOv2 (best: 0.75 vs 0.69), especially at the start (0.57 vs 0.29) - we attribute this to the lack of positional bias. In a sparse, low-label regime where labels are randomly distributed across the image, it is possible for the classifier with DINOv2 to fit to spurious positional information (see \cref{fig:trainable_seg} and \cref{supp:fig:more_trainable_seg}) - this does not happen for ALiBI-Dv2.

\rebuttal{We also included DVT as a comparison. It performs similarly to DINOv2, but slightly worse. }

\subsection{Impact of Encoder}

\begin{figure}
\centering
    \includegraphics[width=\linewidth]{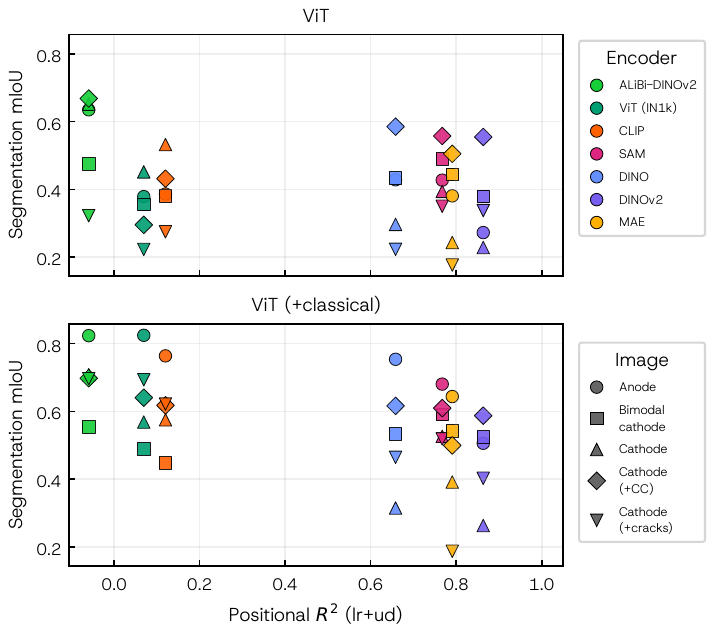}
    \caption{
    \rebuttal{Trainable segmentation mIoU (relative to a ground truth with many labels) using sparse labels vs positional $R^2$ (measured across the microscopy dataset) for a set of different ViT backbones and microscopy images (from \cref{fig:trainable_seg} and \cref{supp:fig:more_trainable_seg}). We compare the results both with and without the concatenation of classical features. In both scenarios, our ALiBi-Dv2 model performs the best due to its combination of reduced positional bias and rich semantic features. }}
    \label{supp:fig:trainable_seg_encoder}
\end{figure}

\rebuttal{In \cref{supp:fig:trainable_seg_encoder}, we examine the impact of the base ViT feature model on trainable segmentation performance in the sparse-label setting.
In order to have a numerical segmentation quality comparison, we produced ground truth segmentations using multiple rounds of labelling on a commercial microstructural segmentation tool (`Polaron').
We then repeated the sparse-label experiment of \cref{sec:trainable_seg} with a set of ViT backends: DINO, DINOv2, ViT (MAE), ViT (In1k), SAM, CLIP, and our ALiBi-Dv2.}

\rebuttal{For each image and backbone, we extracted features, concatenated classical features, and train an XGBoost classifier (default parameters) to map from the features to the same set of sparse labels.
We then compared the mIoU of the predicted segmentation with the ground truth.
This was plotted as the $y-$axis, and for the $x-$axis we used the backbone's average $R^2$ score for the $(x,y)$ linear probe on the micrograph dataset.
We performed this experiment both with and without the concatenation of classical features.
What we observe is that without classical features, the SSL models like DINOv2 (which have richer semantics) perform better than their supervised counterparts.
Once the homogeneous classical features are added, the less positionally biased models perform better.
The intuition here is that the classical features are quite good at solving the problem already, and any improved alignment between the two is useful, whereas any misalignment (from, say, positional biases) become more deleterious.
In all cases, our ALiBi-Dv2 models perform the best - we argue this is because it retains the desirable rich semantics of DINOv2, whilst being less positionally biased.
}
	
\end{document}